%% file: qsnet.tex
\documentclass[acmtog,nonacm]{acmart}
\pdfoutput=1

\usepackage{booktabs} 

\author{Kartic Subr}
\affiliation{
	\institution{University of Edinburgh}
	\country{U.K.}
}

\usepackage{array} 
\usepackage[ruled]{algorithm2e} 
\usepackage{enumitem} %
\usepackage[most]{tcolorbox}
\usepackage{mathtools}
\usepackage{amsmath}
\usepackage{hyphenat}

\SetAlFnt{\small}
\SetAlCapFnt{\small}
\SetAlCapNameFnt{\small}
\SetAlCapHSkip{0pt}

\setcopyright{none}
\makeatletter
\let\@authorsaddresses\@empty
\makeatother

\begin{document}
\title{Q-NET: A Network for Low-dimensional Integrals of Neural Proxies}
\begin{abstract}
Many computer graphics applications require the calculation of integrals of multidimensional functions.
A general and popular procedure is to estimate integrals by averaging multiple evaluations of the function.
Often, each evaluation of the function entails costly computations. The use of a \emph{proxy} or surrogate for the true function is useful if repeated evaluations are necessary. The proxy is even more useful if its integral is known analytically and can be calculated practically. We propose the use of a versatile yet simple class of artificial neural networks ---sigmoidal universal approximators--- as a proxy for functions whose integrals need to be estimated. We design a family of fixed networks, which we call Q-NETs, that operate on parameters of a trained proxy to calculate exact integrals over \emph{any subset of dimensions} of the input domain.
We identify transformations to the input space for which integrals may be recalculated without resampling the integrand or retraining the proxy. We highlight the benefits of this scheme for a few diverse computer graphics applications such as inverse rendering, generation of procedural noise, visualization and simulation.
The proposed proxy is appealing in the following contexts:
the dimensionality is low ($<10$D);
the estimation of integrals needs to be decoupled from the sampling strategy;
sparse, adaptive sampling is used;
marginal functions need to be known in functional form;
or
when powerful Single Instruction Multiple Data/Thread (SIMD/SIMT) pipelines are available for computation.
\end{abstract}

\input{macros}
%
%
%
%
%
%
\maketitle

\graphicspath{{./figs/}}

\input{intro}

\input{relwork}
\input{theory}
\input{results}

\input{cga}

\input{discussion}

\appendix
\section{Appendix}
\subsection{Derivation}\label{sec:derivation}

\input{deriv}

\subsection{Error}\label{sec:error}
The approximation error of shallow feedforward networks~\cite{barron1993universal,shaham2015provable} is bounded by
$|| \f(\x) -\fw(\x) ||_2^2 < \epsilon $ where $\epsilon = \tilde{h}/k^{2c/d}$. $\tilde{h}$ is the first moment of the Fourier spectrum of \f\  which is $c$ times differentiable. Writing $\f_\Delta(\x) \equiv \f(\x)-\fw(\x)$,
\begin{align}
\nonumber
\MoveEqLeft
|| \f(\x) -\fw(\x) ||_2^2 = \DefInt{\Dom}{}{\f_\Delta^2(\x)}{\x} \\
\nonumber
&= \DefInt{\Dom}{}{\f_\Delta^2(x)}{\x}
        \; - \; \left(\DefInt{\Dom}{}{\f_\Delta(\x)}{\x}\right)^2
        \; + \; \left(\DefInt{\Dom}{}{\f_\Delta(\x)}{\x}\right)^2\\
\nonumber
&= \mathrm{V}[\f_\Delta(x)]
    \; + \; \left(\DefInt{\Dom}{}{(\f(\x)-\fw(\x))}{\x}\right)^2 \\
\nonumber
&= \mathrm{V}[\f_\Delta(x)]
    \; + \; (I-\mu)^2
\end{align}
where $\mathrm{V}[.]$ is the variance operator. Substituting this into the bounds for approximation error we have $(I-\mu)^2 < \epsilon - V[\f(\x)-\fw(\x)]$.

\bibliographystyle{ACM-Reference-Format}
\bibliography{qsnet}

\end{document}

%% file: macros.tex
\newtcolorbox{myframe}[1][]{colback=yellow!80!gray!10!white,colframe=white,#1}

\newcommand{\figref} [1]{Fig.~\ref{fig:#1}}
\newcommand{\secref} [1]{Sec.~\ref{sec:#1}}
\newcommand{\eqnref} [1]{Eq.~\ref{eq:#1}}
\newcommand{\hdg} [1]{\vspace{.5em}\noindent\textbf{#1}}

\newcommand{\myfigfull}[4]{%
	\begin{figure*}[#1]%
		\centering%
		{#2}%
 		\caption{\label{fig:#4} #3}%
		\Description{#3}%
	\end{figure*}%
}

\newcommand{\myfig}[4]{%
	\begin{figure}[#1]%
		\centering%
		{#2}%
 		\caption{\label{fig:#4} #3}%
		\Description{#3}%
	\end{figure}%
}
\newcommand{\DefInt}[4] {\ensuremath{\displaystyle\int\limits_{#1}^{#2} #3 \; \mathrm{d}{#4}}}
\newcommand{\Exp}[1]{\ensuremath{\left<#1\right>}}
\def\one{\mbox{1\hspace{-3.85pt}\fontsize{11}{14.4}\selectfont\textrm{1}}} 

\newcommand{\RR}{\ensuremath{\mathbb{R}}} 
\newcommand{\Dom} {\ensuremath{\mathcal{D}}}
\newcommand{\sig} {\ensuremath{\sigma}}
\newcommand{\f} {\ensuremath{f}}
\newcommand{\wa} {\ensuremath{W_1}}
\newcommand{\waij} [2]{\ensuremath{W_1^{#1,#2}}}
\newcommand{\wbv} {\ensuremath{\w_2}}
\newcommand{\wav} {\ensuremath{\w_1}}
\newcommand{\wb} {\ensuremath{\w_2}}
\newcommand{\ba} {\ensuremath{\mathbf{b}_1}}
\newcommand{\bb} {\ensuremath{b_2}}
\newcommand{\x} {\ensuremath{\mathbf{x}}}
\newcommand{\xsi} {\ensuremath{\mathbf{x}_n}}
\newcommand{\Sp} {\ensuremath{\Gamma}}
\newcommand{\Pl} [1]{\ensuremath{\mathrm{Li}_{#1}}}
\newcommand{\p} {\ensuremath{{S}}}
\newcommand{\w} {\ensuremath{\mathbf{w}}}
\newcommand{\fw} {\ensuremath{f_{\w}}}
\newcommand{\hw} {\ensuremath{h_{\w}}}
\newcommand{\gw} {\ensuremath{g_{\w}}}
\newcommand{\y} {\ensuremath{\mathbf{y}_i}}
\newcommand{\ty} {\ensuremath{\tilde{\mathbf{y}}_i}}
\newcommand{\uu} {\ensuremath{\mathbf{u}}}
\newcommand{\q} {\ensuremath{q}}
\newcommand{\wq} {\ensuremath{\p}}
\newcommand{\sigq} {\ensuremath{\sigma_q}}
\newcommand{\bq} {\ensuremath{\ba^i}}
\newcommand{\fslw} {\ensuremath{\tilde{f}_{\w}}}
\newcommand{\wasl} {\ensuremath{\tilde{W}_1}}
\newcommand{\basl} {\ensuremath{\tilde{\mathbf{b}}_1}}

\newcommand{\T} {\ensuremath{\theta}}
\newcommand{\Lt} {\ensuremath{\ell}}
\newcommand{\Lo} {\ensuremath{\ell_o}}

%% file: intro.tex
\myfigfull{htbp!}{
\begin{tabular}{@{}b{.8\linewidth}b{.2\linewidth}@{}}
    \raisebox{-.45\height}{\includegraphics[width=\linewidth] {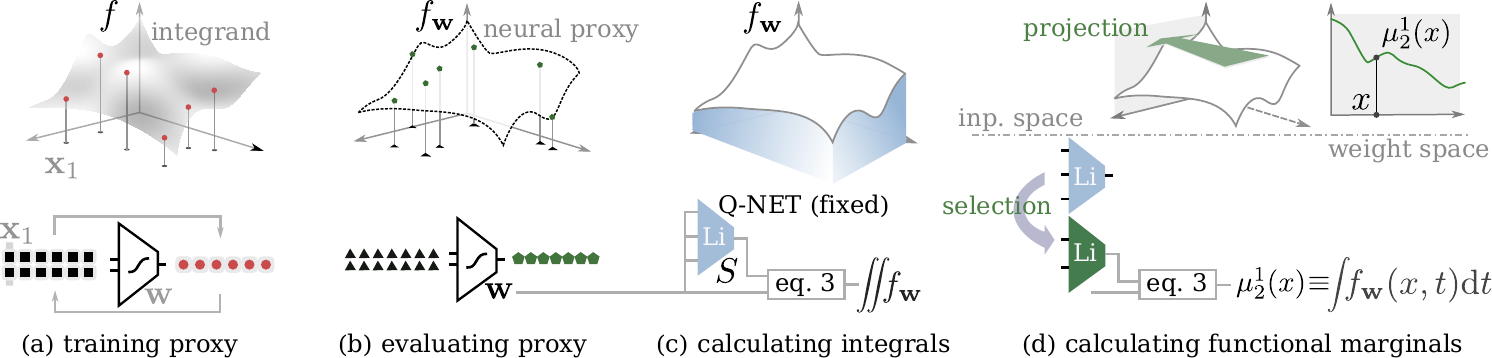}}&
    \resizebox{.185\textwidth}{!}{%
        \begin{tabular}{@{}rl@{}}
        INP. SPACE & WT. SPACE\\
        \vspace{.2em} \\
        affine trans. &  matrix mult.\\
        summing & concatentation\\
        slicing & column select\\
        projection & selection\\ 
        sub-domain & modify $S$\\        
        \vspace{.63em} \\
        \multicolumn{2}{c}{(e) other operators} 
        \end{tabular}
    }
\end{tabular}
}
{Overview. (a) Given samples of an unknown function \f\ and (b) a 1-layer sigmoidal universal approximator trained with these samples to regress a proxy function \fw, we derive a formula in terms of weights \w\ to calculate the integral of \fw. (c) We design a fixed network (no learnable parameters) which we call a Q-NET as an elegant mechanism to perform calculations. (d) Any marginal (projection) of \fw\ can be represented in  functional form via an input selection transformation to Q-NETs. (e) A list of other transformations to \fw\ which can be accommodated via modification of network parameters.} {overview} 

\section{Introduction} \label{sec:intro}




The estimation of integrals is a computational bottleneck across computer graphics applications such as rendering, simulation of dynamics and visualization. The \textit{integrand}---or function whose integral is sought---is rarely available in closed form and is potentially costly to evaluate. ~e.g.~it might require tracing rays, performing intersection tests or finding nearest neighbors. An obvious strategy is to replace integrands with surrogates or \emph{proxy} functions that are easier to evaluate. Another use of the proxy is as an interpolant for sparsely sampled data. The choice of a suitable proxy is important, especially for applications involving multidimensional discontinuous functions when sampling and reconstruction~\cite{MDAS08,Koschier2017} are challenging. Since numerical integration is considered more sample-efficient than reconstruction~\cite{durand2011frequency,SubrKautz13,VisSamp12}, a faithful proxy is also expected to be effective for numerical integration provided that a procedure is known to calculate its integral. 

Artificial neural networks are versatile representations of functions that have led to groundbreaking results in a variety of learning problems such as reconstruction (regression), classification and dimensionality reduction. Deep neural networks are able to approximate difficult functions (high-dimensional, with discontinuities, etc.). While numerous works have tailored neural network architectures to various applications, estimating integrals of functions learned by these architectures remains an open problem.
We investigate a specific, well known class of shallow feed-forward networks (SFFN) that consists of one hidden layer with a sigmoid activation function and a purely linear output layer. 
This textbook case is an example of a \emph{universal approximator network}~\cite{cybenko1989approximation,hornik1991approximation,lu2017expressive} and it can approximate~\cite{shaham2015provable} and integrate~\cite{NNI20} any continuous function $\f: \RR^d \rightarrow \RR$ accurately. During training, samples of \f\ are used to learn parameters \w\ of the network so that it represents the functional approximation $\fw\approx\f$ which can be evaluated rapidly anywhere in the domain. 
The simplicity and universality of sigmoidal approximators make them an attractive first choice as neural proxies for numerical integration. Despite their lack of sophistication, they are surprisingly practical for a variety of applications (see ~\secref{apps}). 

Given a trained proxy \fw, we design a family of shallow neural networks (Q-NETs) to evaluate the exact functional marginals of \fw\ in closed form by integrating over any subset of the input space $\RR^d$. Approximate functional projections of \f\ can therefore be obtained using \w\ without further sampling of \f. The proxy \fw\ is also useful (\secref{experiments}) as a control variate~\cite[Sec.~8.9]{mcbook}.
We derive fixed weights to define Q-NETs (\secref{qnet}) for simplifying integral calculations with interpretable properties. We also derive transformations to Q-NETs or their inputs to accommodate operations such as projection (see \figref{overview}) which allows integrals of transformed functions to be calculated without resampling \f\ or retraining \fw. Our approach is practical since standard implementations of SFFN may be leveraged for fast computation on graphics processing units.


\hdg{Contributions:}
\begin{itemize} [leftmargin=1.3ex,nosep,labelsep=.4em,itemindent=0em,parsep=.4em]
	\item we derive a fixed-weight (no learnable parameters) SFFN with one hidden layer to calculate integrals and obtain marginal functionals of functions represented by sigmoidal approximators (\secref{qnet}) ; 
	\item we derive transformations to the inputs and network weights to act as counterparts to operations on the input space (\secref{ops});
	\item we perform quantitative tests to assess the empirical fidelity of integration via Q-NETs in the presence of discontinuities (\secref{experiments}) ;
	\item we present qualitative examples of the proxy's versatility by applying it to a variety of computer graphics applications such as inverse rendering, modeling, visualization and simulation (\secref{apps}).
\end{itemize}

%% file: relwork.tex
\section{Related work} \label{sec:relwork}
Functions with multiple jump discontinuities (see \figref{vizintegr}c.) occur commonly in CG applications.~e.g.~shadows on a checkerboard. The general procedure to estimate integrals of such functions using point samples remains inefficient even in as few as ten dimensions.  Many practical adaptations have been developed for computer graphics applications by introducing domain-specific assumptions. 

\hdg{Numerical integration (general)} Quadrature schemes~\cite{brassquadrature,DesignedQuad18} approximate definite integrals by dividing the integration domain into cells (usually uniformly along each dimension), approximating the function using polynomials within each cell and summing up the analytically computed integrals within each cell. As the dimensionality of the domain increases, the number of cells and hence number of samples required increases exponentially. This is referred to as the `curse of dimensionality'.
Monte Carlo (MC), Quasi-MC (QMC) and Markov Chain MC (MCMC) operate differently, by expressing integrals as expectations which can be estimated via simulation~\cite{MetUlamMC}. The simulation is (pseduo-) random for MC and MCMC. Although these methods converge slowly at $O(1/\sqrt{N})$, they escape the curse of dimensionality. Several variants~\cite{mcbook} address the slow convergence by striking a compromise between \textit{bias} (accuracy) and \textit{variance} (precision). QMC methods~\cite{niederreiter1978,QMCBook} replace stochasticity with carefully designed, deterministic samples which improves convergence dramatically when integrands are smooth and integration is over moderate dimensionalities.
 
\hdg{Computer Graphics adaptations} Quadrature schemes have been used for antialiasing~\cite{GTumblin96}, to render participating media~\cite{HTexPerlin89,GQuadTangled11,PVR17}, for discretized time-integration of Laplacians physics-based animation~\cite{kharevych2006geometric} and subspace deformation~\cite{AKD08Cubature}. MC (or QMC) path tracing~\cite{RE86} and its MCMC variants~\cite{VeachThesis98} form  the industry standard~\cite{PTProd2019Sig,christensen16path} for estimating multidimensional integrals in offline rendering applications. MC and MCMC methods have been honed for rendering via analyses in Fourier~\cite{FALT05,CovTracing5D}, wavelet~\cite{AWR09} and gradient-domains~\cite{lehtinen2013gradient,GDPT15}. Recently, a neural control variate~\cite{NCV20} that was tailored to light transport estimation produced an impressive reduction in variance. The discretization of time for physics based animation necessitates a different class of integrators~\cite{PBASig19,HOTI20} which reformulate differential systems variationally and solve time integration as an optimization problem. 

\hdg{Machine Learning applications} 
Bayesian methods routinely use probabilistic model-based surrogates to improve sample-efficiency for expensive integrands~\cite{BMC03} or to guide active sampling (adaptive Bayesian quadrature~\cite{PereiraABQ12,ConvBQ19}, Bayesian Optimization~\cite{BOReview}, etc.). This approach can be used  to estimate the marginal likelihood~\cite{BQ15,FBQ14}, approximate the posterior~\cite{BALP15,AGP18} and to simultaneously infer both~\cite{Acerbi2018}.  They operate by imposing a Gaussian Process (GP) prior on the integrand and using analytical formulae for the expectation and uncertainty of the statistical surrogate. 
Hybrid probabilistic neural networks use GPs to model neural weights~\cite{karaletsos2020hierarchical} for calibrated reasoning about uncertainty.  \textit{Integral representations}~\cite{dereventsov2019neural} use an analysis of continuous distributions, for a particular target function, from which instances of shallow neural networks can be sampled. Deep neural networks can mimic MC solutions to certain partial differential equations~\cite{grohs2019deep}, sidestepping the curse of dimensionality. 

\hdg{Integration using Neural Networks}
Integration formulae have been derived for shallow networks with different activation functions~\cite{NINN06,Spline313} in 1D. Recent work proves the existence of integration formulae in closed form for functions represented by sigmoidal SFFN in polyhedral domains~\cite{NNI20} and derives formulae for hyperrectangular domains. We review their  formula as background (~\secref{prelims}) and present the derivation in an appendix. Another class of methods ~\cite{Teichert_2019,Song20} operates by treating the neural network function as the integral and train its derivatives to match the integrand. Once trained, the neural network is used to evaluate the integral. While these methods enable the use of deeper networks, it is not obvious how they may be used to evaluate multiple marginals once trained. 
 

%% file: theory.tex
\myfig{htbp!}
{
\includegraphics[width=\linewidth]{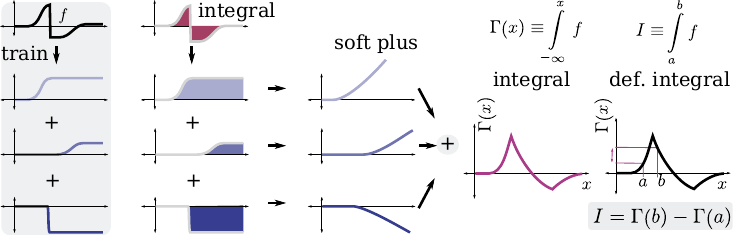}
}
{\emph{Insight}: A sum of sigmoids (trained network) can be integrated analytically as a weighted sum of their respective integrals (soft plus in 1D). We extend this to multiple dimensions and simplify its calculation. }{insight}

\myfigfull{!htbp}{
	\includegraphics[width=\linewidth]{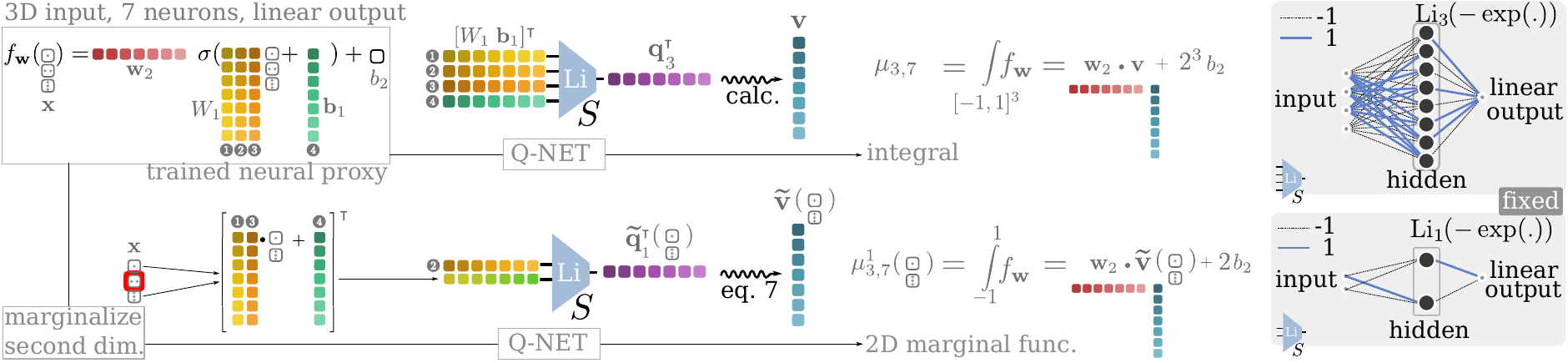}
}
{Given weights $\w \equiv \left(\wa, \wb, \ba, \bb \right)$ of a sigmoidal approximator for a function in 3D, the integral of the function (top) and its projection (bottom) are calculated similarly but with different instances of Q-NETs (blue trapezoids). In the latter case, the 2D marginal is a function of the first and third input variables. Q-NETs have fixed weights (rightmost) and are reconfigurable based on the number of dimensions being integrated. }
{qnetillus}

%


\section{Theory} \label{sec:theory}
The key insight to integrate sigmoidal approximators is that the integral of a sum of shifted and scaled sigmoids is a weighted sum of the integrals of the sigmoids (see ~\figref{insight} and the submitted video). The sigmoid and its integrals are instances of a special function called the polylogarithm~\cite{Eulerbook,lewin1981polylogarithms} which can be evaluated efficiently~\cite{crandall2006note}. The definite integral over a $d$-dimensional hyperrectangle is obtained by accumulating integrals at its vertices with appropriate signs.~e.g.~if $\Gamma$ is a 2D cdf, the definite integral over $[0,1]\times[0,1]$ is given by $\Gamma(1,1) - \Gamma(1,0) - \Gamma(0,1) + \Gamma(0,0)$. 
We simplify its evaluation, enable calculation of marginals and avoid re-sampling \f\ and retraining \fw\ for certain transformations.  

\subsection{Notation and background} \label{sec:prelims}
We denote row and column vectors with boldface characters (e.g.~\x, \ba, \wb) and matrices using capital letters (e.g.~\wa). We use superscripts to select elements of a vector or matrix.~e.g.~ $\ba^i$ and $\wa^{i,.}$ represent the $i^{th}$ element  \ba\ and $i^{th}$ row of \wa\ respectively.
Without loss of generality (see \secref{discussion}) we assume a normalised hyper-rectangular domain $\x \in \Dom \equiv \left[-1,1\right]^d$. We approximate the function $\f:\Dom \rightarrow \RR$ with $\fw:[-1,1]^d\rightarrow [-1,1]$ obtained by a training a shallow feedforward neural network with one hidden layer ($k$ neurons) and a linear output layer using $N$ samples ${\f(\x_n)}\;n=1,\cdots,N$. \w\ collectively encodes all learnable parameters of the network: a $k\times d$ matrix \wa, a $k$-dimensional row vector \wb, a $k$-dimensional column vector \ba\ and a real number \bb\ so that 
\begin{equation}        
	\label{eq:ffnet}
	\fw (\x) = \wb  \;\; \sig(\wa\x + \ba )+\bb, \; \mathrm{where} \;\;
	\sig^i(\mathbf z^i) \equiv \frac 1 {1+e^{-\mathbf z^i}}
\end{equation}  
operates independently on each of the $k$ elements. 
We design a network with fixed weights to calculate $\mu_{d,k} (\w)$, the integral of \fw\ over $d$-dimensional hyper-rectangular domains \Dom, in closed-form. $\mu_{d,k} (\w)$ is a consistent\footnote{Since we choose $k \propto N^{\frac{1}{1+d}}$. It is not consistent in practice, for a fixed $k$.} estimator of the integral of \f:   
\begin{eqnarray}
\label{eq:probstmt}
\mu_{d,k} (\w) = \DefInt {\Dom} {} {\fw(\x)} {\x} \; \approx \; I
 & \mathrm{where } &
I\equiv\DefInt{\Dom}{}{\f(\x)}{\x}. 
\end{eqnarray}

\subsubsection{Formula}\label{sec:formula}
The formula\footnote{\noindent The derivation of the formula is shown in  \secref{derivation}. It was concurrently and independently derived in an earlier version of this manuscript (submission to NeurIPS 2020) and by Lloyd et al~\shortcite{NNI20}.
} for the integral of \fw\ is:
\begin{align}
\label{eq:qnetd}
	\mu_{d,k}(\w) &= {\wb} \; \mathbf{v} \; + \; 2^d \; b_2, \; \\
\nonumber
    \mathrm{where}  \; \mathbf{v}^i 
                    &=	2^d  \;+\; 
                        \frac 1  {\tilde{\w}_1^i }  
	 		\; \displaystyle\sum\limits_{m=1}^{2^d} 
			{\alpha_m \; \Pl d \left( -\exp (\p^{m,.}\; \wa^{i,. \;\top}  - \ba^i)\right)}.	
\end{align}
Here $\mathbf v$ is a column vector representing integrals of each of the neurons in the hidden layer and $\Pl d (x)$ is the polylogarithm function of order $d$. The summation is due to the definite integral requiring appropriate addition or subtraction of the integral at each of the vertices of the hypercube. The $2^d$ vertices (rows) are represented by $\p$, whose elements are $\pm1$. The contribution at each vertex is positive ($\alpha_m=1$) when there are an even number of $-1$s in the row $\p^{m,:}$ and negative ($\alpha_m=-1$) otherwise. The division by $\tilde{\w}_1^i\equiv\prod_j \wa^{i,j}$, the product of the elements of the $i^{th}$ row of \wa, arises due to the integration of transformed sigmoids.

\subsubsection{Affine transformations} \label{sec:affine}
If the input space is transformed as $ M \x + c$ where $M$ is a transformation matrix and $c$ is a translation, 
\begin{eqnarray}	
    \nonumber
	\fslw(\x) & = &  \wb  \;\; \sig\left(\wa \; (M \; \x + c) + \;  \ba \right)+\bb \\
	\label{eq:fftr}
                  & = &  \wb  \;\; \sig(\wasl \; \x \; + \;  \basl )+\bb, 
\end{eqnarray}  
where $\wasl \equiv \wa M $ and $\basl \equiv \ba + \wa c$.

\subsubsection{Sums of integrands} Since \sig\ is applied element-wise to its vector, the terms in a linear decomposition of \f\ may be trained separately. That is, if 
$\f(\x) = \lambda_g \, g(\x) + \lambda_h \,h(\x)$ and  \gw\ and \hw\ are  independently trained proxies for $g(\x)$ and $h(\x)$ respectively, then:
\begin{equation}
	\label{eq:sumproxy}
	\fw (\x) = [\lambda_g\wb^g  \;  \; \lambda_h\wb^h]  \;\; \sig\left(
								\begin{bmatrix}\wa^g \\ \wa^h \end{bmatrix} \x + 
								\begin{bmatrix}\ba^g \\ \ba^h \end{bmatrix}
								\right) \;\; +\;\; \lambda_g\bb^g \;+\; \lambda_h\bb^h.
\end{equation}	
As an exception, we use superscripts here to denote the function used to train the weights. 
 Sums of functions that are trained independently may be achieved using a single proxy with a concatenation of weights as in~\eqnref{sumproxy}. 
 This is a consequence of having only one hidden layer and the output layer being linear. This property simplifies distribution of training effort when \f\ can be decomposed as a sum.

\subsection{Q-NETs}\label{sec:qnet}
Our central observation is that rewriting the elements of $\mathbf{v}$ in \eqnref{qnetd} as 
$\mathbf{v}^i = 2^d + q_d(\y)/\tilde{\w}_1^i$ allows it to be written in terms of: 
\begin{myframe}[floatplacement=h]
\begin{align}
    \label{eq:qnetq}    
    q_d(\y)  &\equiv  \w_3 \;\sigq \left( \left[ \p \; \;-\one_{2^d}\right] \y \right) \\
    \nonumber 
    \y &\equiv [ \wa^{i,.\;} \; \ba^i ] ^{\top}.    
\end{align}
\end{myframe}
\noindent Here $\sigq(.) \equiv \Pl{d} (-\exp(.))$, and $\one_{2^d}$ is a column of $2^d$ ones. The advantage of this representation is that \eqnref{qnetq}\ is similar in structure to \eqnref{ffnet} and can therefore be computed using a feedforward network with fixed weights (no learnable parameters), input $\y$, output $q_d(\y)$, one hidden layer containing $2^d$ neurons and the activation function $\sigq$. The biases of this network are zero and its input and output weights are $\left[ \p \; \;-\one_{2^d}\right]$ and $\w_3$ respectively. $\w_3$ is a row vector whose $m^{th}$ element is $\alpha_m$. We call this family of networks, parameterized by $d$, Q-NETs since it enables quadrature of the approximator network.  
In practice, all $k$ vectors may be stacked (as columns) into a matrix $\left[\wa \; \ba\right]^\intercal$ for efficient (vectorized) evaluation. In this case $\w_3$ will be replaced by a matrix each of whose $k$ rows is a vector $\w_3$.  See the top row of \figref{qnetillus} for an example of the role of a Q-NET in calculating integrals.
Q-NETs as defined above form an elegant representation with interpretable properties. 

\subsection{Operations on Q-NETs} \label{sec:ops}
We highlight key properties (also see the accompanying video) of Q-NETs which we exploit in the applications shown in~\secref{apps}.

\subsubsection{Affine transformation} The integral of the transformed function $\fslw(\x)$ may be calculated just as for \f, but with the modified weights $[ \wasl^{i,.\;} \; \basl^i ] ^{\top}$ in~\eqnref{fftr} as inputs to the Q-NET and then divided by the absolute value of the determinant of the Jacobian of the affine transformation. Using this, integrals over non-axis-aligned hyperrectangles may be computed, as in our example that estimates optical depth along rays (\secref{optdepth}). 

\subsubsection{Projection} \label{sec:projection}
To marginalize $r<d$ input dimensions of \x, the variables $\x^{1,\cdots, r}$ need to be integrated\footnote{We assume without loss of generalization that the first $r$ dimensions are marginalized.}, yielding a function in the remaining variables $\mu^r_{d,k}(\w, \x^{r+1,\cdots, d})$. Using Q-NETs, the procedure is similar to the full integral but with $\tilde{\mathbf{v}}^i$ instead of $\mathbf{v}^i$ where 
\begin{eqnarray}
	\label{eq:qtovmargin}
        \tilde{\mathbf{v}}^i \;\; &\equiv& \;\; 2^r \;+\; { q_r(\ty)} \;\;/\;\; {\prod\limits_{j=1}^{r}\wa^{i,j}}.
\end{eqnarray}
Compared to \eqnref{qnetq} $d$ has been replaced with $r$ on the rhs and the input to the Q-NET is $\ty \equiv [ \wa^{i,1,\cdots, r} \;  \; (\ba^i + \wa^{i,r+1,\cdots, d} \x^{r+1,\cdots,d}) ] ^{\top} $  instead of \y. Thus projection along a subspace of \x\ amounts to a \emph{selection} operation in \y\ where the weighted non-marginalized variables are moved from being individual inputs to Q-NET to an aggregate input along with  the last (bias) dimension of \y. 

\subsubsection{Slicing}
If the proxy is sliced through $r<d$ dimensions using the constants $\x^{1,\cdots, r} = c^{1,\cdots, r}$, the resulting $(d-r)$ dim. function is
\begin{equation}
	\label{eq:ffsl}
	\fslw(\x^{r+1,\cdots, d}) \; = \;  \wb  \;\; \sig\left(\wasl \; \x^{r+1,\cdots, d} \; + \;  \basl \right)+\bb,
\end{equation}  
where $\wasl \equiv \wa^{.,(r+1, \cdots, d)}$ and $\basl \equiv \ba + \wa^{.,(1, \cdots, r)} c^{1, \cdots, r}$. That is, slicing along subdimensions of \x\ amounts to removing the corresponding columns of \wa\ and adding the product of those columns with the slicing constants to the bias \ba. Again, since the sliced function can be obtained via manipulation of the original weights, the integral of the sliced function may be calculated just as before via a Q-NET but with the following two changes: 1) the dimensionality of integration is $d-r$ instead of $d$; and 2) the input to the Q-NET is $[ \wasl^{i,.\;} \; \basl^i ] ^{\top}$ with the weight matrix and bias vector from~\eqnref{ffsl}.

\subsubsection{Integrals over sub-domains}
Our choice to work in a normalized domain $[-1,1]^d$ manifests in the Q-NET formulation in two ways: first, the constant $2^d$ is the product of the differences between upper and lower limits in each dimension; and second, matrix \p\ contains $\pm1$ to represent the vertices of the hyperrectangle. To modify the integration limits to $\mathbf{a},\mathbf{b}\in[-1,1]^d$, $2^d$ needs to be replaced by $\prod (\mathbf{b}^j-\mathbf{a}^j)$ in \eqnref{qnetd} and \eqnref{qnetq} ($j$ iterates through $d$ dimensions). Next, the rows of \p\ must be updated with combinations of elements of $\mathbf{a}$ and $\mathbf{b}$ to list vertices of the hypercube defined by the new limits.

\myfig{!htbp}{ 
		\begin{tabular}{@{}c@{}c@{}c@{}} 
		\includegraphics[width=.326\linewidth]{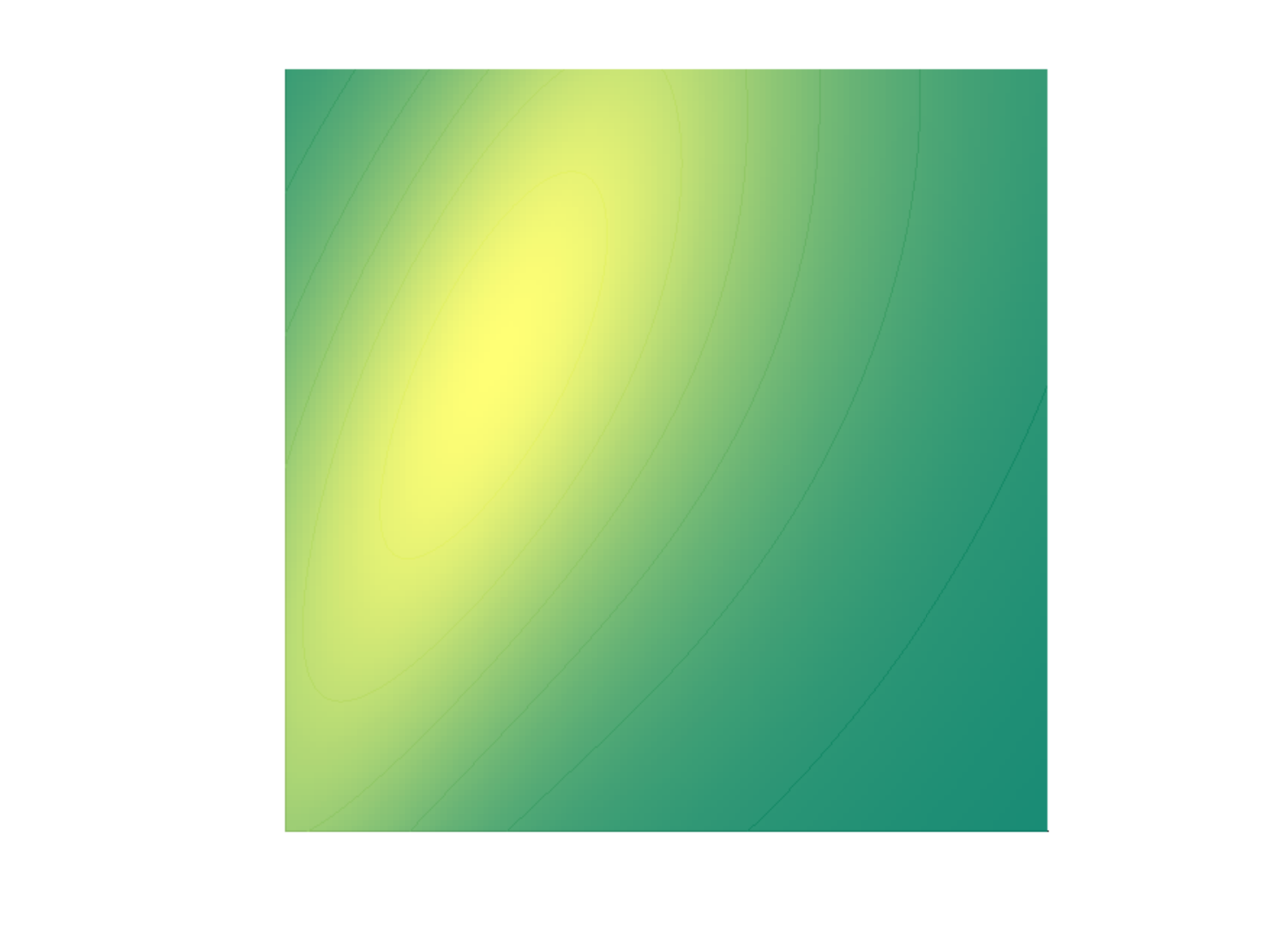}&
		\includegraphics[width=.326\linewidth]{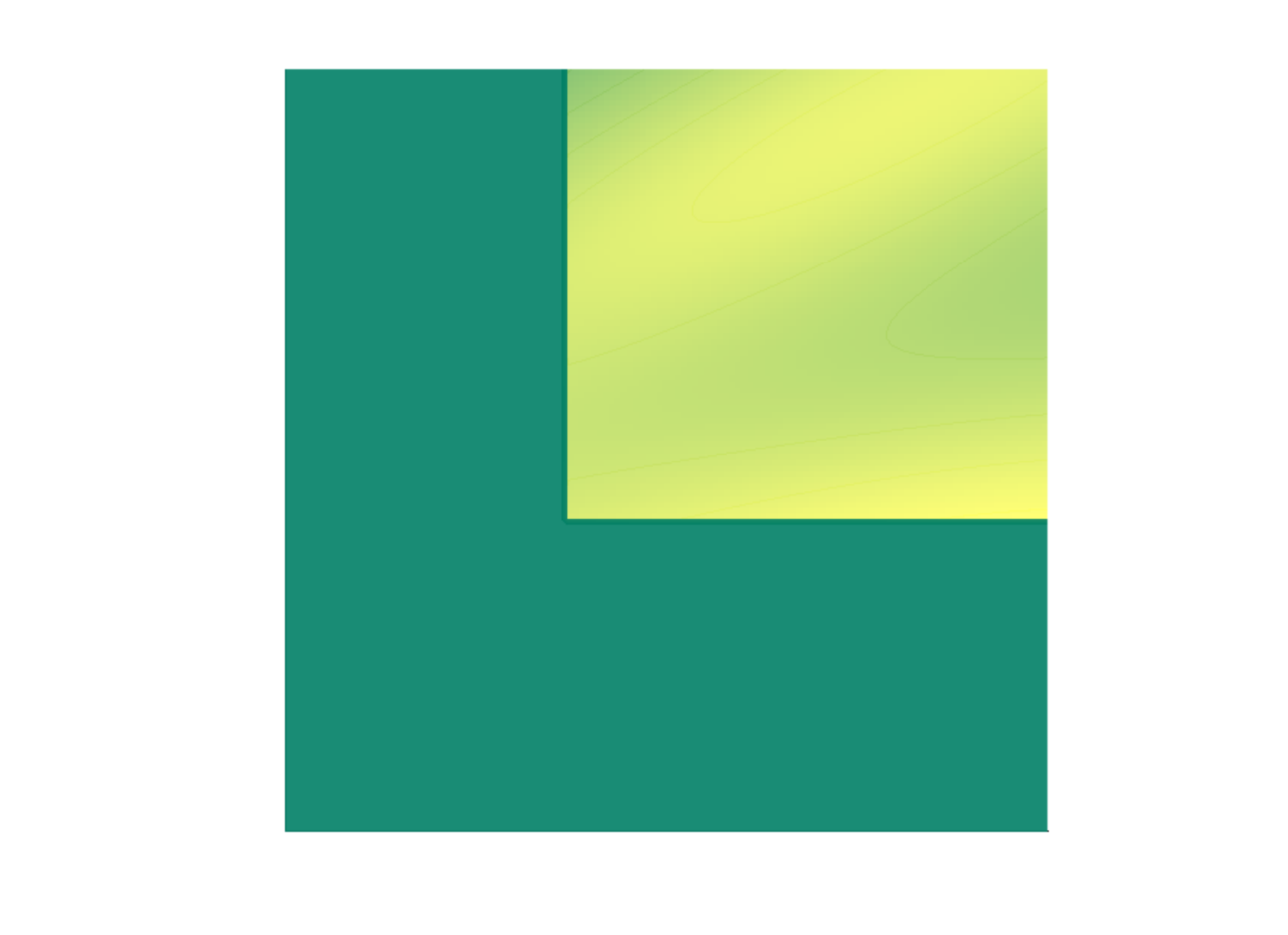}&
		\includegraphics[width=.326\linewidth]{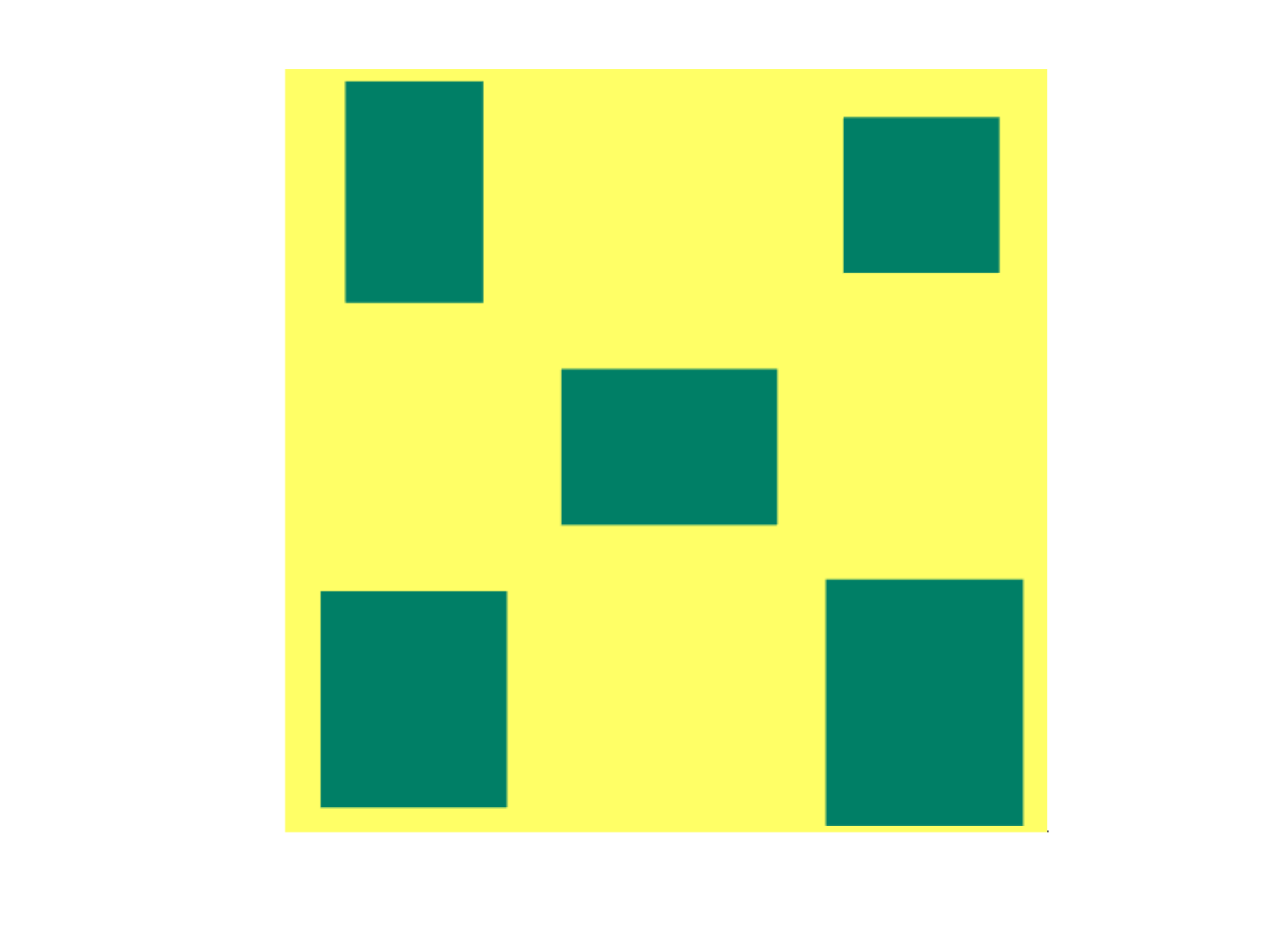} \\
		\includegraphics[width=.326\linewidth]{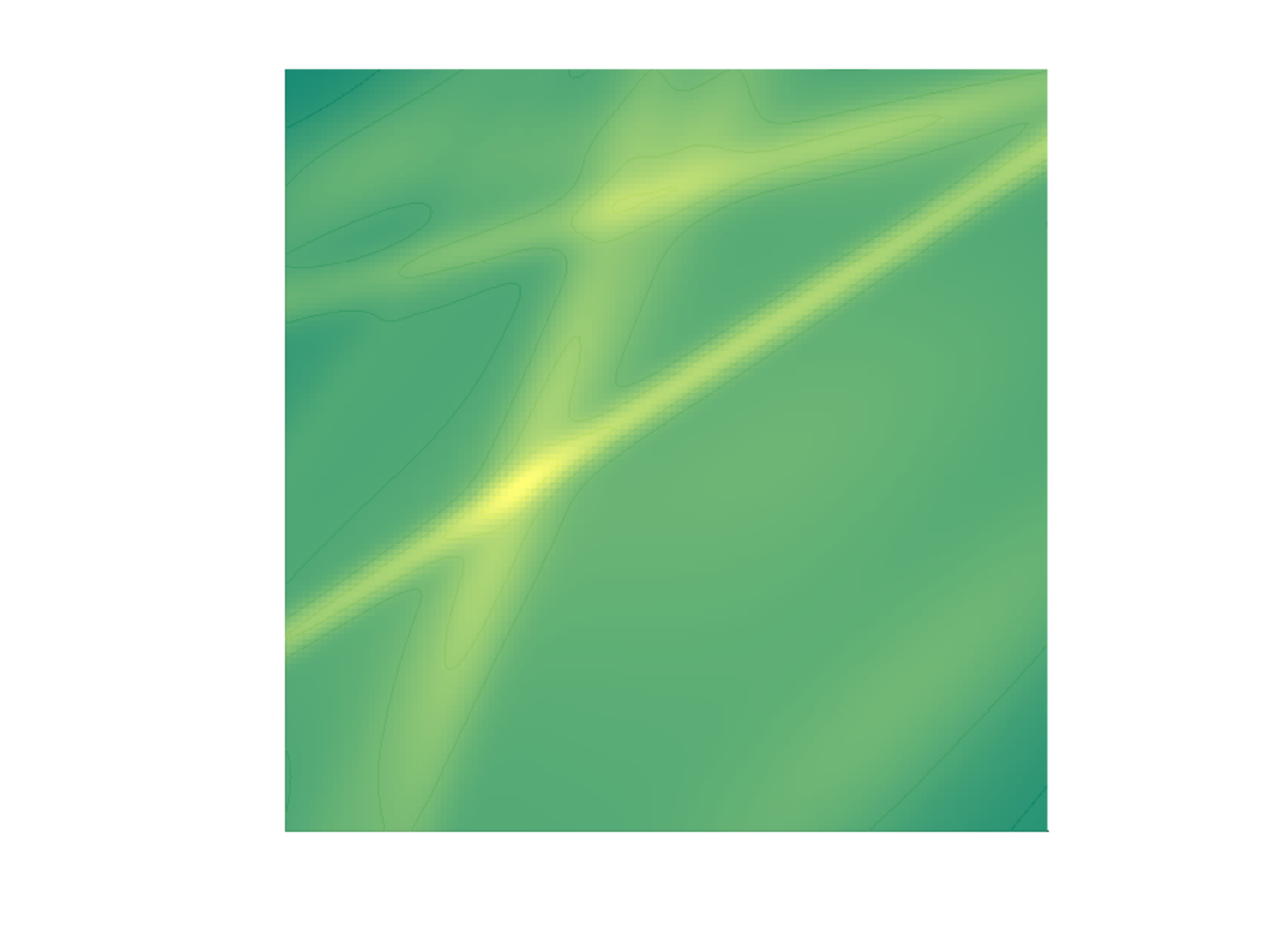}&
		\includegraphics[width=.326\linewidth]{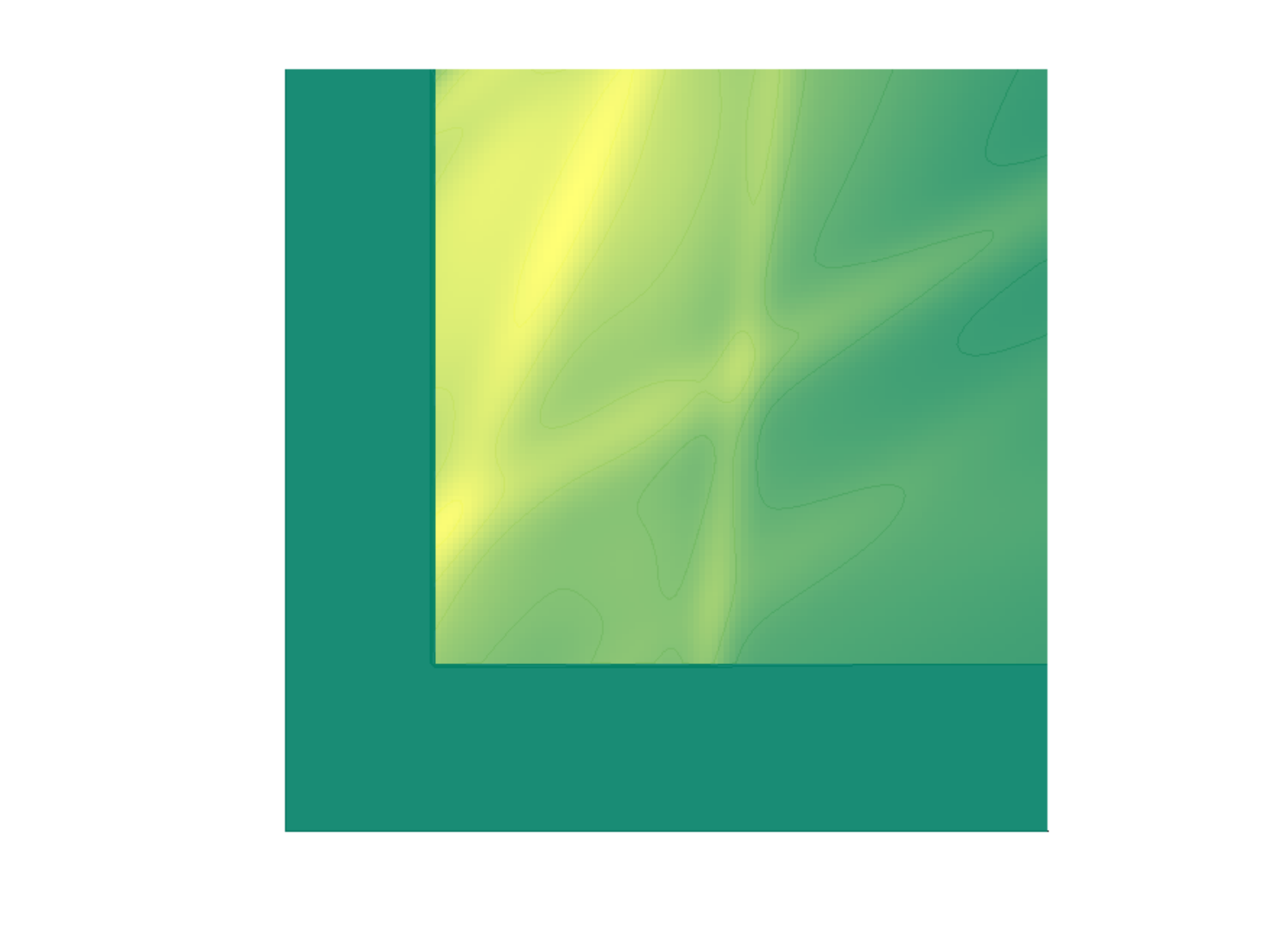}&
		\includegraphics[width=.326\linewidth]{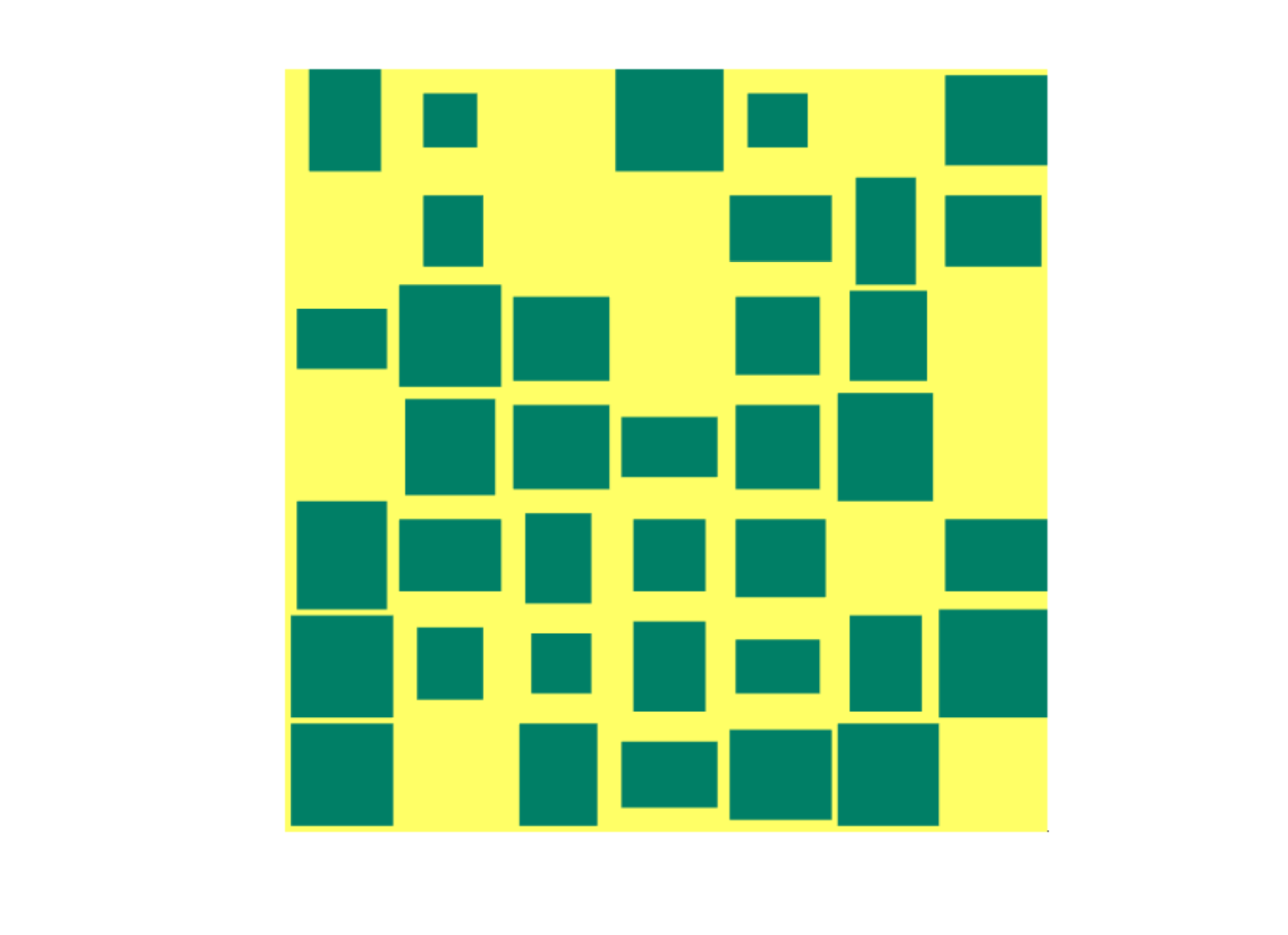}\\
 		a) GM &
 		b) DGM &
 		c) HR
 		\end{tabular}
}{Test integrands for empirical validation: (a) Gaussian mixtures (GM), (b) GM with $d$ discontinuities (GMD) and (c) binary hyperrectangles (HR). Two different parametric settings are shown (rows) for each family.}{vizintegr}

\subsection{Complexity and error}
The computational complexity of evaluating the formula directly is $O( k\; d \; 2^d )$. Using a Q-NET facilitates parallel and vectorized computation across neurons, across dimensions or the $2^d$ rows of the sign matrix as necessary. Given a trained proxy, computation time is independent of $N$, the number of samples. However, the memory complexity is $O(kd)$ for direct evaluation (no need to store \p) compared to $O(kd+d^2)$ with Q-NETs via binary encoding of \p\ which occupies $O(d)$ space. Since $d$ is the integrated dimensions, marginalizing along a small number of dimensions of a high-dimensional function remains feasible.

\begin{theorem}
        \label{th:error}
	The upper bound for the squared error between integrals $I$ and $\mu$ of a $c$-times differentiable function $\f:\RR^d\rightarrow\RR$ and its shallow sigmoidal approximant \fw\ (with $k$ neurons) is given by
	$$
	(I-\mu)^2 \;\; <  \;\; \frac{\tilde{h}}{k^{2c/d}} \;\; - \;\;  \mathrm{V}[\f-\fw].
	$$ 
	$\mathrm{V}[.]$ denotes variance and $\tilde{h}$ is the first moment of \f's Fourier spectrum. 
\end{theorem}
\noindent The bound in Theorem~\ref{th:error} is tighter than that for reconstruction error (see \secref{error}) by the variance term. This is qualitatively consistent with previous analyses of numerical integration of bandlimited signals~\cite{durand2011frequency,SubrKautz13,VisSamp12} which also conclude that integration is relatively more sample-efficient. Since our theorem assumes smoothness, we run experimental validation on integrands (see \figref{vizintegr}) with discontinuities.  

%
%

%% file: results.tex
\newcommand{\Fv} {\ensuremath{\vec{F}}}
\newcommand{\Ss} {\ensuremath{\mathcal{S}}}

\section{Results I: Experimental validation} \label{sec:experiments}
We using three classes of integrands for empirical tests: Smooth integrands represented by Gaussian mixtures (GM), smooth with $d$ discontinuities by GM with discontinuities (GMD) and discontinuous integrands using sums of binary hyperrectangles (HR). We averaged results for different random parameters of these functions in the domain $[0\times 1]^d$ for dimensions $d\leq 12$. \figref{vizintegr} visualizes each class (columns) in 2D for two different choices of parameters (rows).
We ran experiments in MATLAB on a Desktop with an Intel 8-Core i7-6700 processor, 32 GB of RAM and an NVIDIA TITAN RTX GPU. Training times depended on \f, $d$ and $k$, but integration using Q-NETs is typically steady at rates of about 1000, 550 and 300 neurons/second per thread for 1D, 2D, and 3D integrals respectively.

\myfig{h}{
\includegraphics[trim= 145 0 0 0, clip=true,width=\linewidth]{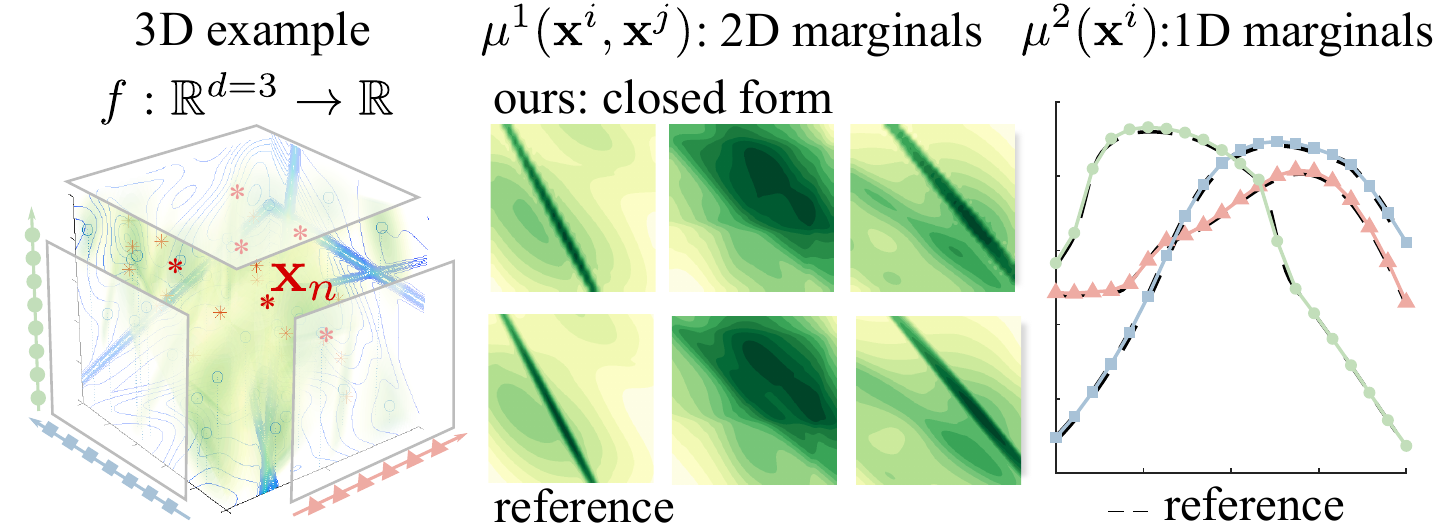}
}{Results of marginalizing a 3D Gaussian mixture (left) along one dimension (middle) and two dimensions (right). The resulting 2D and 1D marginals are evaluated on grids and compared with reference values.}{marginals}

\myfig{!htbp}{
	\begin{tabular}{@{}c@{\hspace{.5em}}c@{}}
		\includegraphics[width=.37\linewidth]{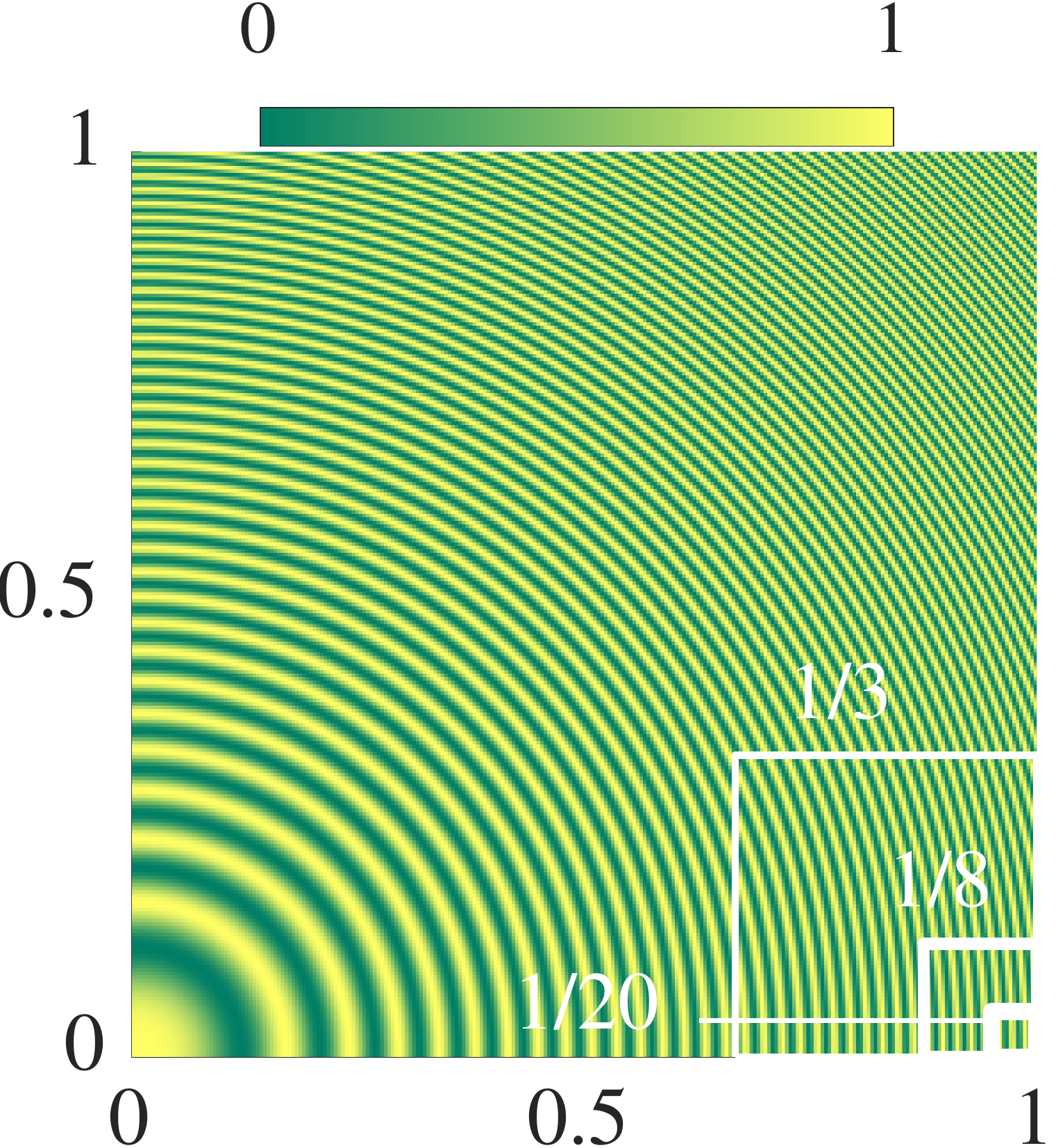}&
		\includegraphics[width=.6\linewidth]{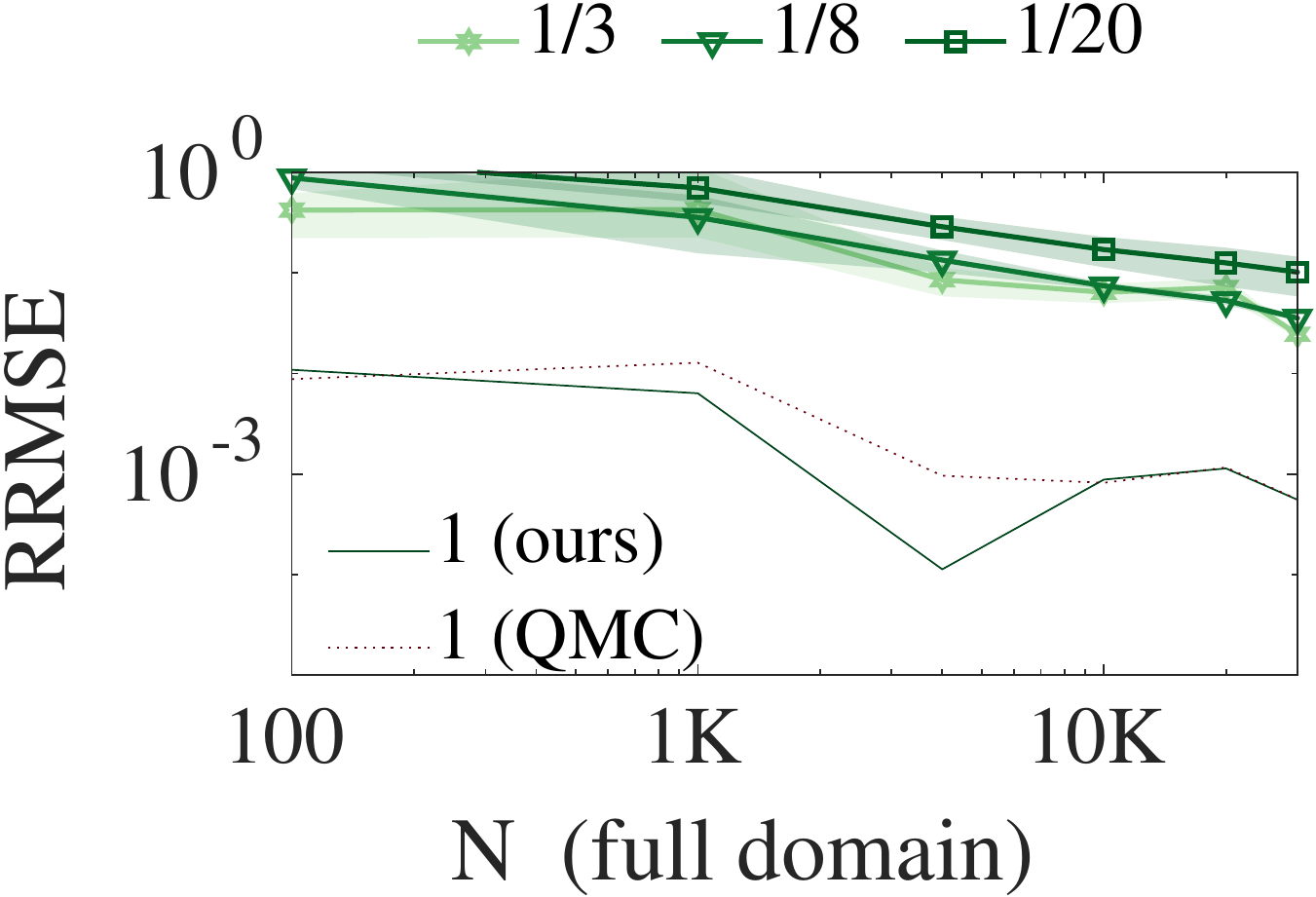}
	\end{tabular}
}{We trained \fw\ on the zone plate function $ (1 + \cos{ (220 x^2 + 220 y^2)})/2$ (left) with $N$ samples in the unit square and tested our method for calculating integrals in sub-domains. The plot shows RRMSE (green curves) of 100 random, square subdomains  of sizes $1/3$, $1/8$ and $1/20$ as $N$ is increased. The errors for integrating over the whole domain is shown (black).}{subint}

\myfig{!htbp}{
\begin{tabular}{@{}m{.41\linewidth}@{\hspace{.4em}}m{.41\linewidth}@{\hspace{.4em}}m{.14\linewidth}@{}}
    \begin{tabular}{c} \includegraphics[width=\linewidth]{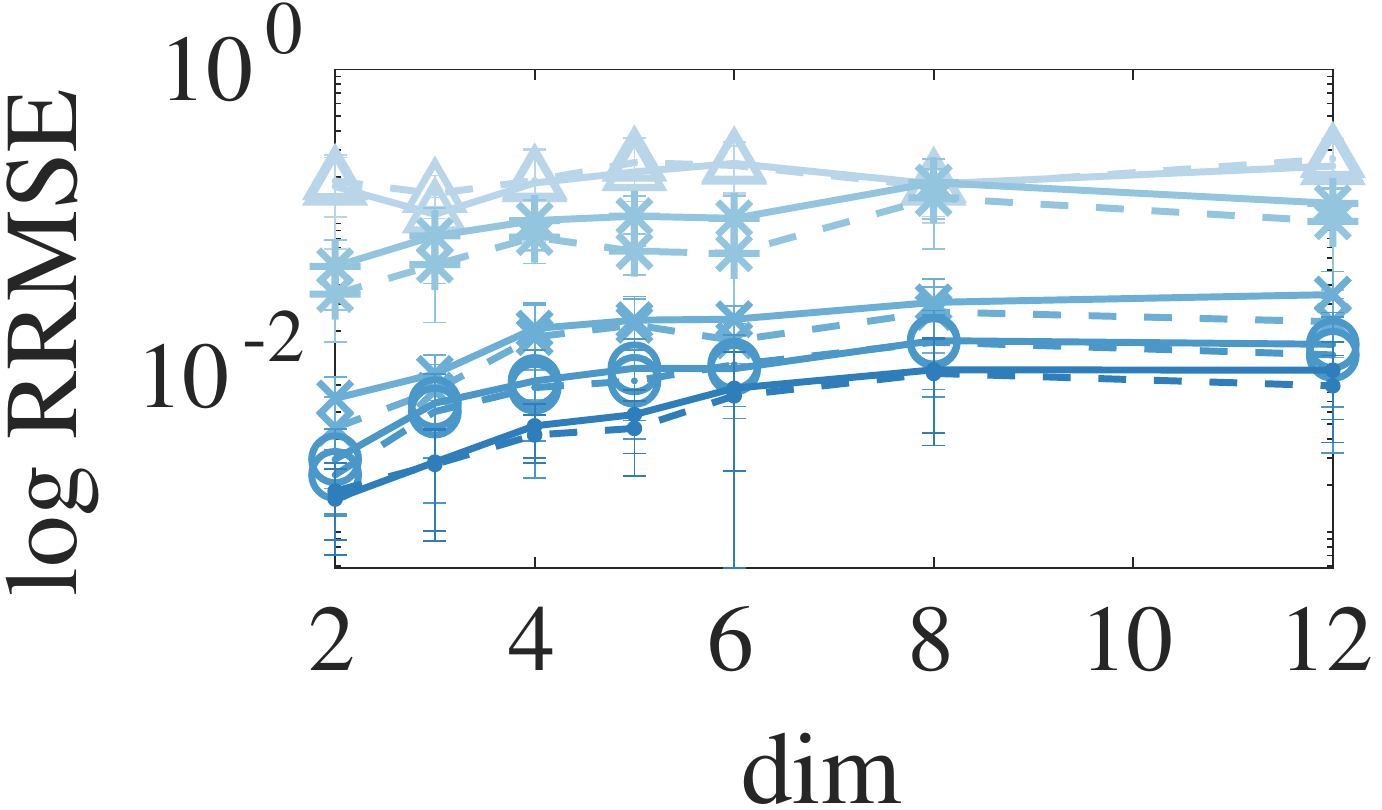} \end{tabular}&
    \begin{tabular}{c} \includegraphics[width=\linewidth]{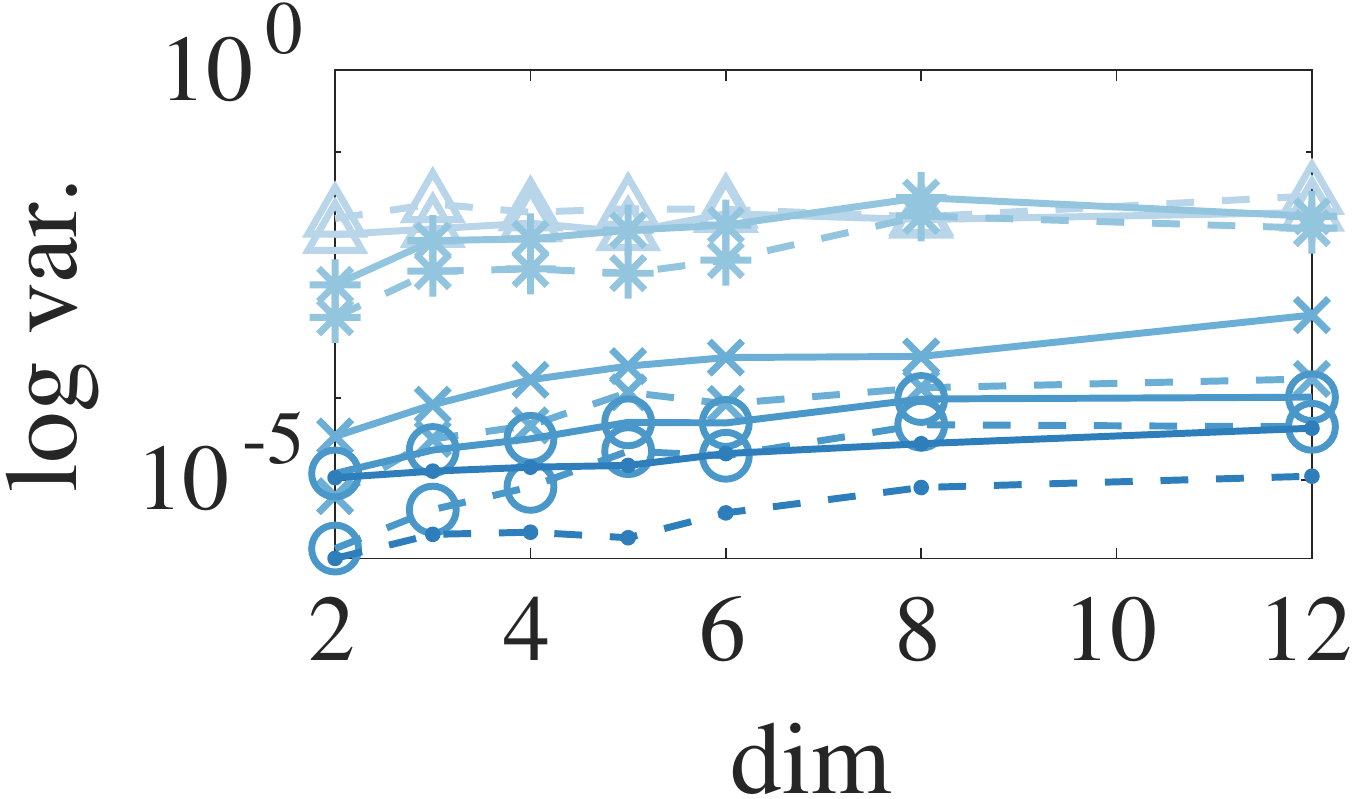}\end{tabular}&
    \begin{tabular}{c} \includegraphics[width=\linewidth]{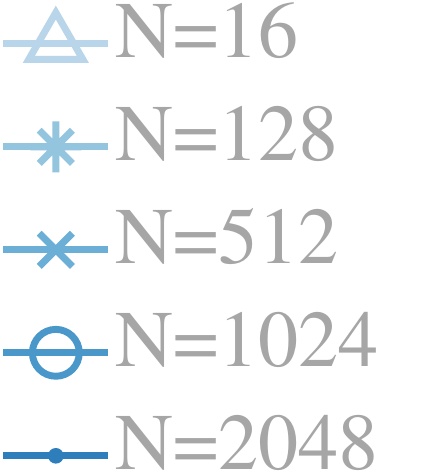}\end{tabular}
\end{tabular}
}{Increasing dimensionality on HR integrands. Plots of relative error (left) and relative variance (right) averaged over 40 iterations each of 50 random hyperrectangle integrands over dimensions $d=2,3,4,5,6,8,12$. Although error increases with dimension, the effect is not sustained.}{dimsfig}

\hdg{Direct use for integration}
For each integrand \f, generated with randomly chosen parameters, we trained \fw\ using $N$ samples. Then we evaluated the formula via a Q-NET and measured the error and variance of the integral by comparing with  analytically calculated references. \figref{rmseintegr} plots the mean convergence of relative root mean squared error (RRMSE) for each class (columns) of integrands for $d=2$ (top row) and $d=5$ (bottom row).  Trend lines (dashed) on the log-log plots depict $O(N^{-0.5})$ and $O(N^{-1.5})$ rates of convergence while error bars indicate standard deviation across $100$ integrands.

\myfigfull{!htbp}{
\scalebox{.9}{
\begin{tabular}{@{}cccc@{}}
        \begin{tabular}{@{}c@{}} \rotatebox{90}{\textbf{2D}} \end{tabular} &
        \begin{tabular}{@{}c@{}}\includegraphics[width=.3\linewidth]{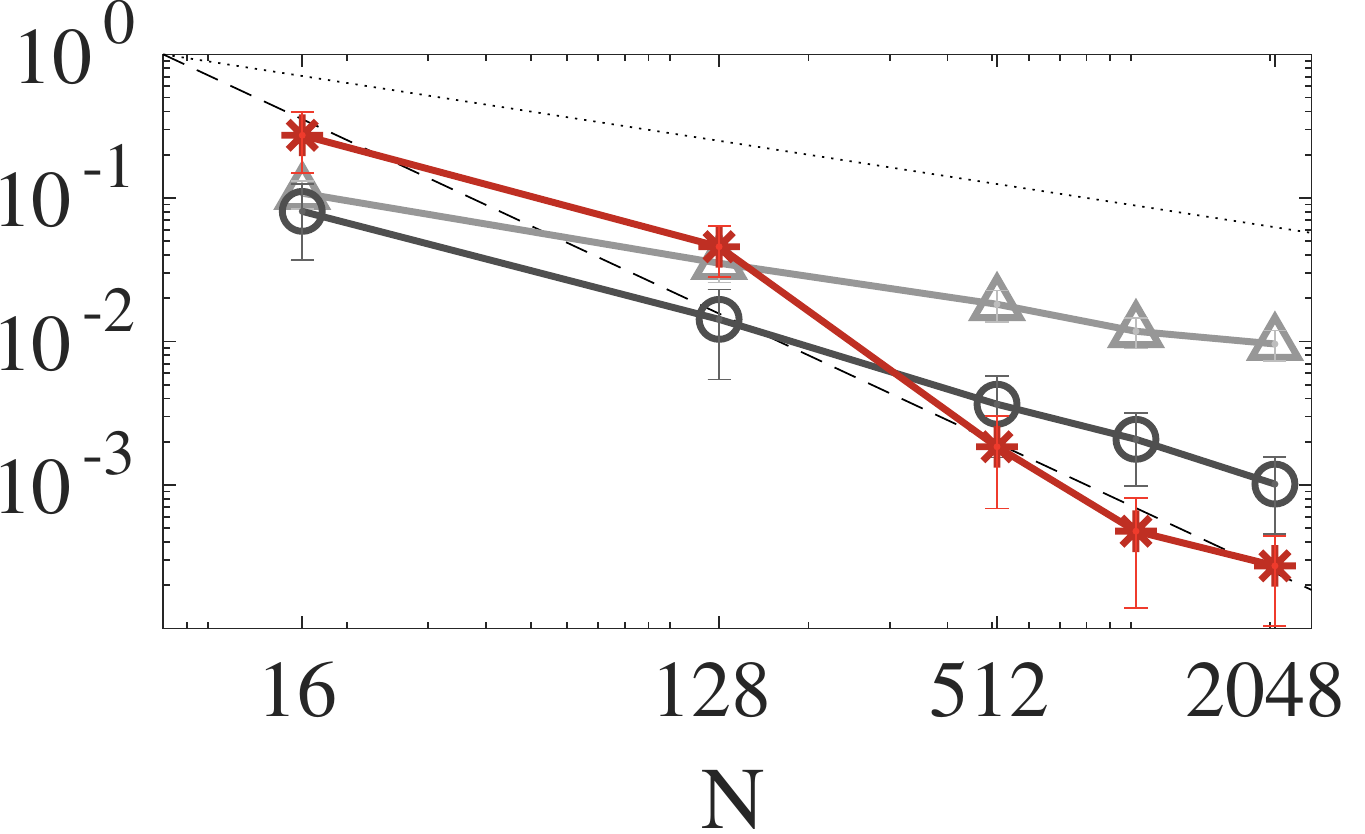} \end{tabular}&
        \begin{tabular}{@{}c@{}}\includegraphics[width=.3\linewidth]{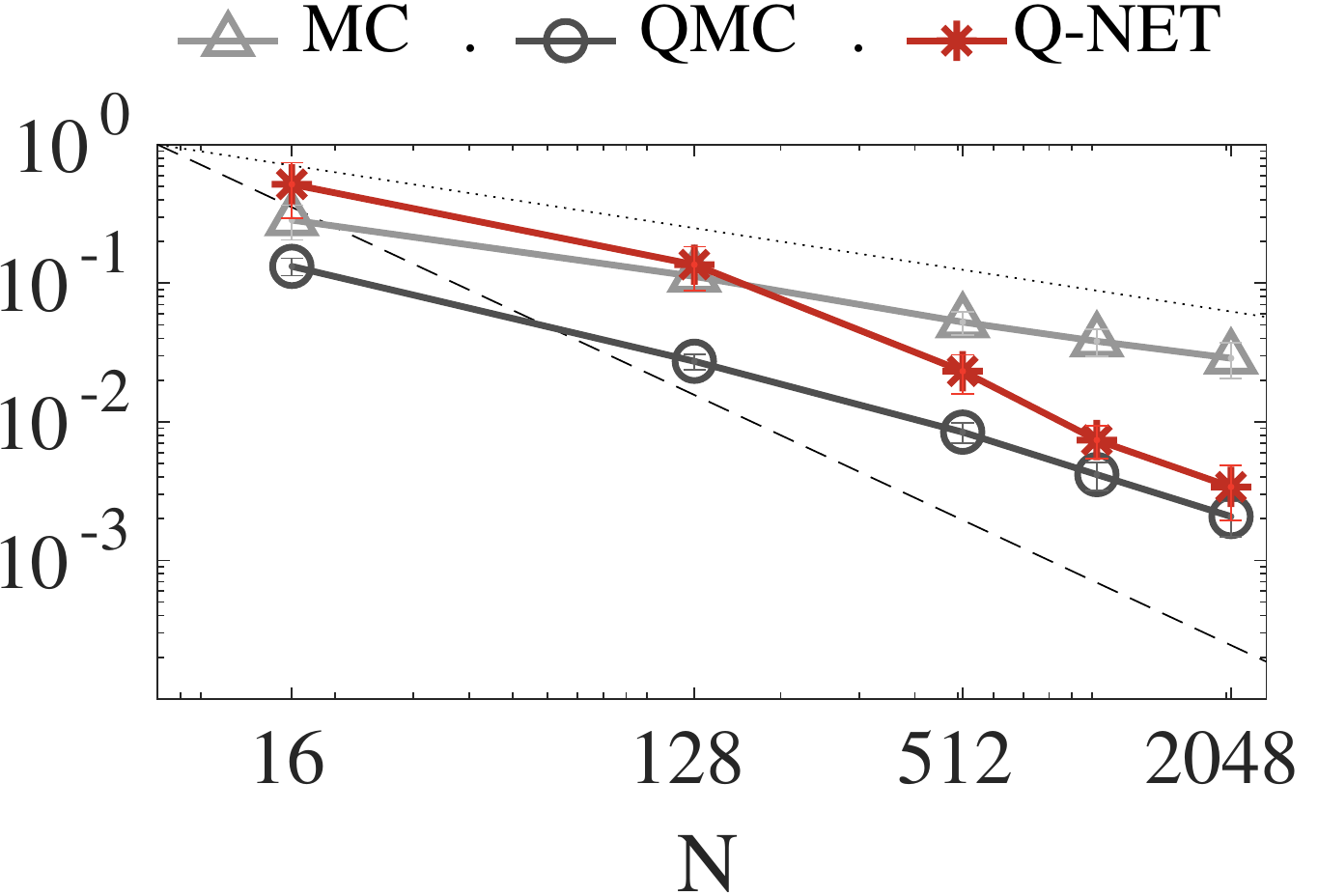} \end{tabular}&
        \begin{tabular}{@{}c@{}}\includegraphics[width=.3\linewidth]{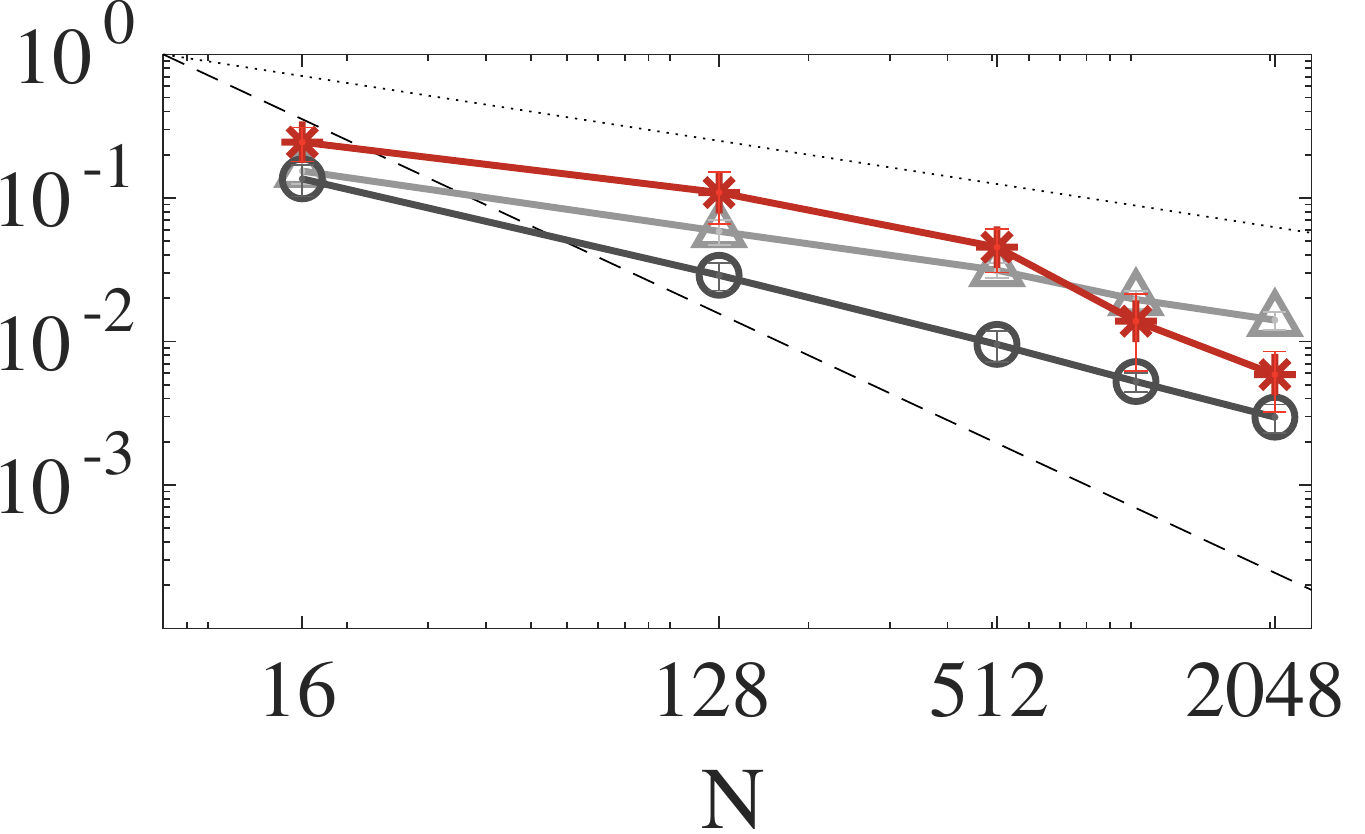} \end{tabular}\\
        \begin{tabular}{@{}c@{}} \rotatebox{90}{\textbf{5D}} \end{tabular} &
        \begin{tabular}{@{}c@{}}\includegraphics[width=.3\linewidth]{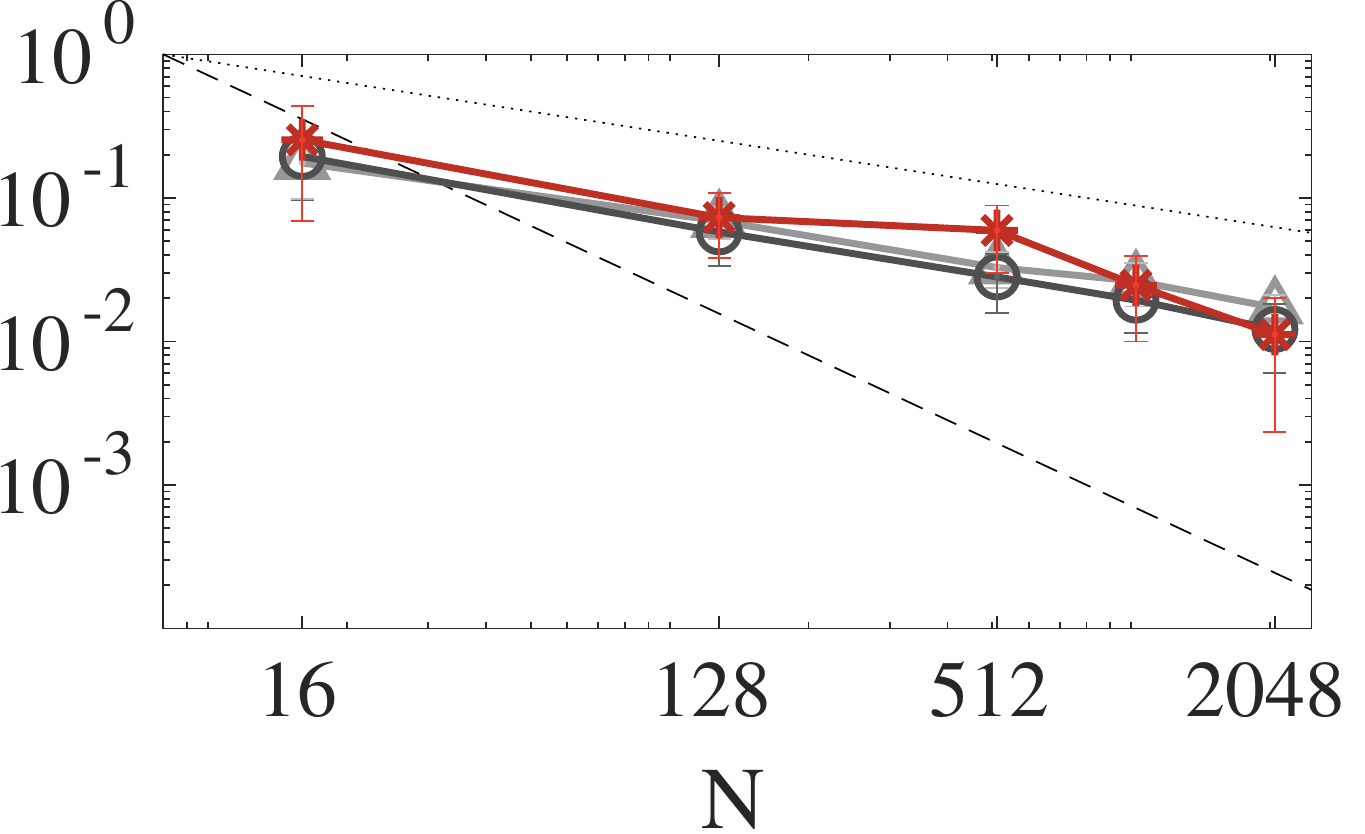} \end{tabular}&
        \begin{tabular}{@{}c@{}}\includegraphics[width=.3\linewidth]{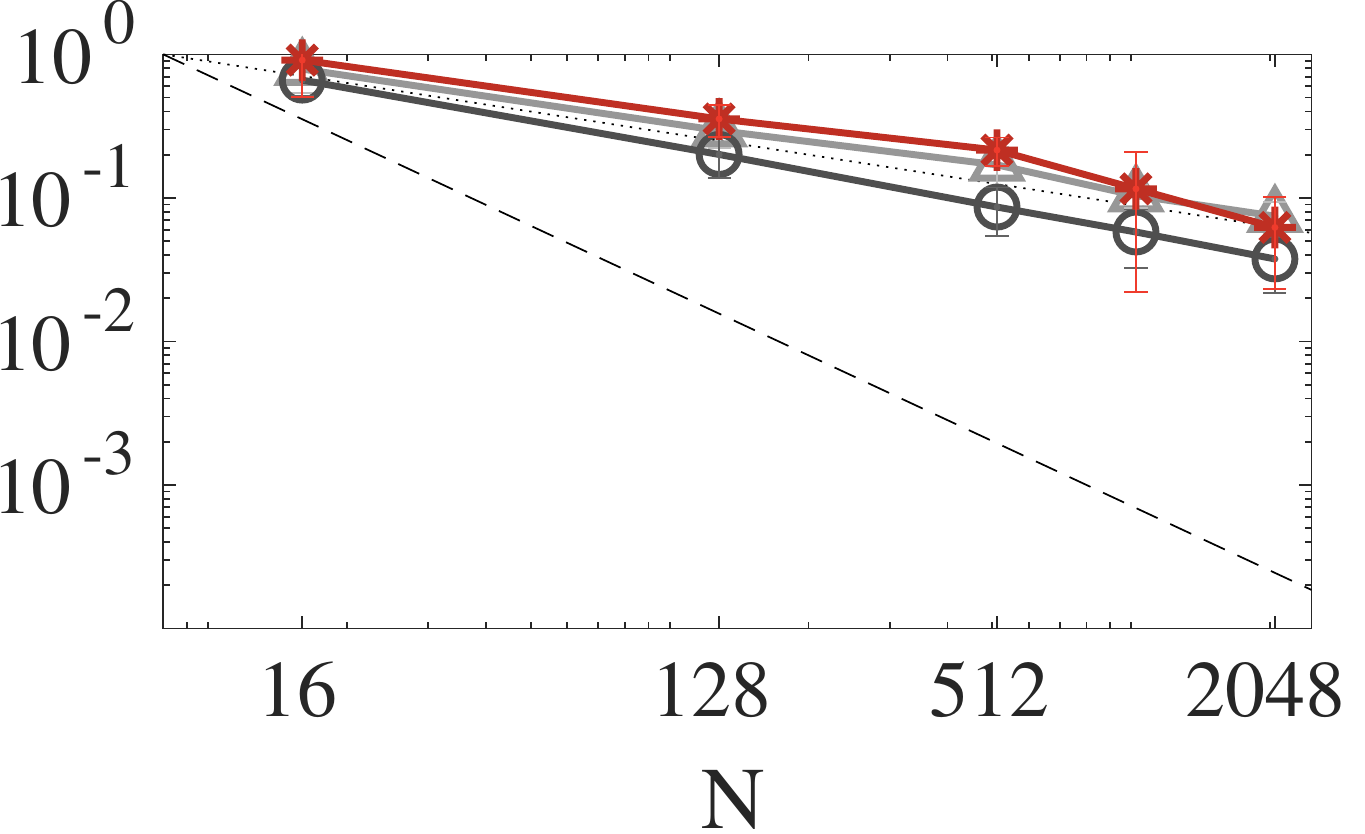} \end{tabular}&
        \begin{tabular}{@{}c@{}}\includegraphics[width=.3\linewidth]{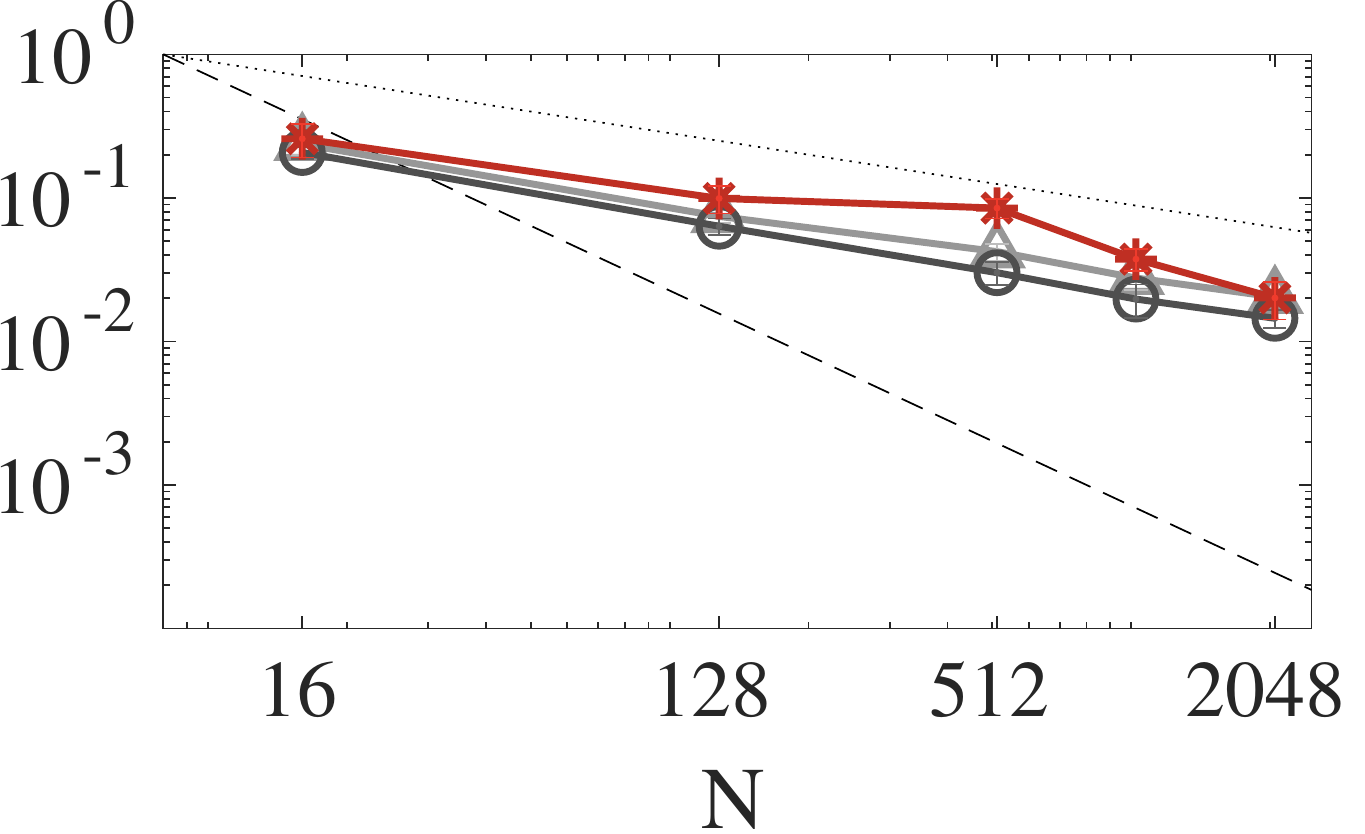} \end{tabular}\\
        & (a) GM & (b) DGM & (c) HR \\
\end{tabular}
}
}{Plots of relative error of the integral of a neural proxy vs number of samples used to train the proxy for different classes of integrands (columns). The plots show empirically that Q-NETs yield a consistent estimator. Relative errors of MC and QMC estimators are shown for comparison.}{rmseintegr}

\hdg{Increasing dimensionality}
The plots in \figref{rmseintegr} confirm that our estimate (red) is consistent, improves upon general Monte Carlo integration and is competitive with Quasi-Monte Carlo methods in 2D. However in 5D the proxy seems less effective when discontinuities are present (e.g.~HR). We investigated this further by measuring relative errors and variances for dimensions up to $d=12$ using HR integrands. \figref{dimsfig} plots these for different $N$ (colors) and two choices for $k$, the number of neurons (solid and dashed lines) using a log scale for error and a linear scale for dimensions. Although these plots confirm the increase in error from $d=2$ to $d=5$, they also provide reassurance that the increase flattens down towards $d=12$.

\hdg{Use as a control variate}
We devised a family of estimators CV-Q-NET, parametrized by  $\nu\in\left[0,1\right]$, that use \fw\ as a control variate~\cite[Sec.~8.9]{mcbook} to integrate \f. Given $\nu$, CV-Q-NET uses $\mathrm{ceil}[(1-\nu)N]$ samples to train \fw\ and the remaining samples to integrate $\f_\Delta (\x) = \f(\x)-\fw(\x)$ via standard MC (or QMC). The final estimator is then the sum of the closed-form integral of \fw\ and the MC (or QMC) estimator. When $\nu=0$, CV-Q-NET is equivalent to $\mu$ (Q-NET) and as $\nu$ tends to one it approaches pure MC (or QMC). As predicted by theory, while this does not improve error in~\figref{cverrvar}, it results in variance reduction for a wide range of $\nu$.

\myfigfull{!htbp}{
		\begin{tabular}{@{}cc@{\hspace{3 ex}}cc@{}}
		\includegraphics[width=.195\linewidth]{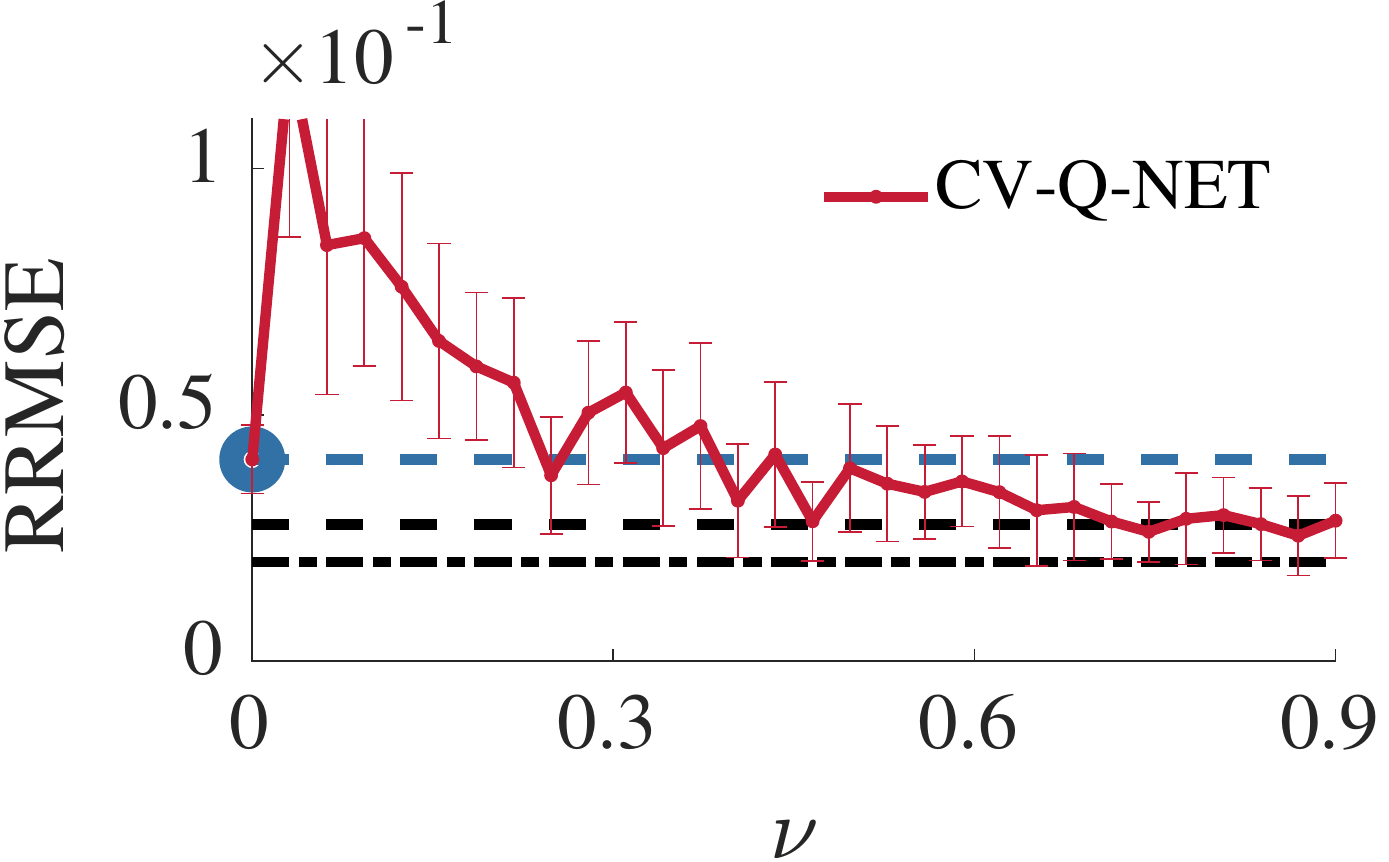}&
		\includegraphics[width=.195\linewidth]{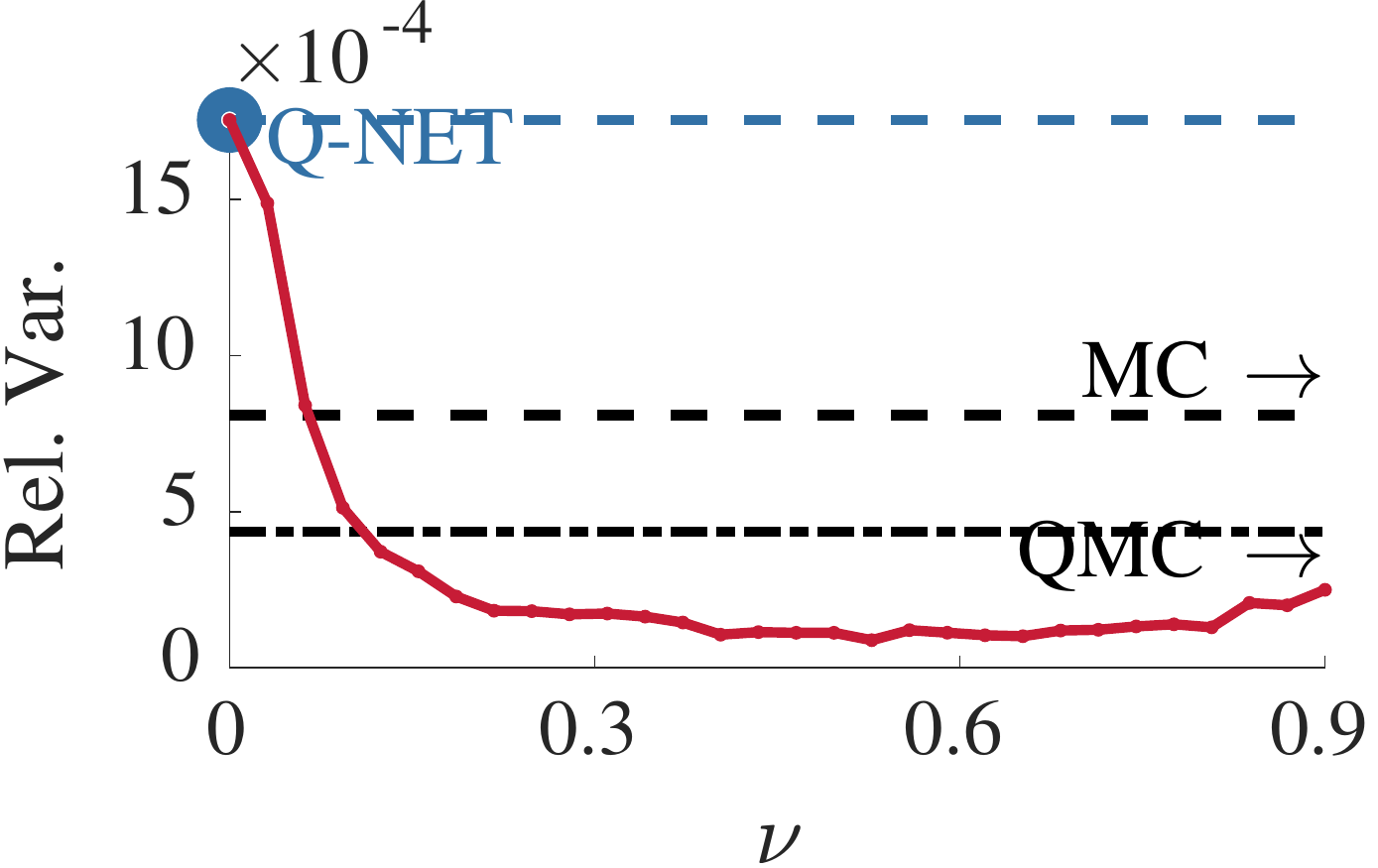}&
		\includegraphics[width=.195\linewidth]{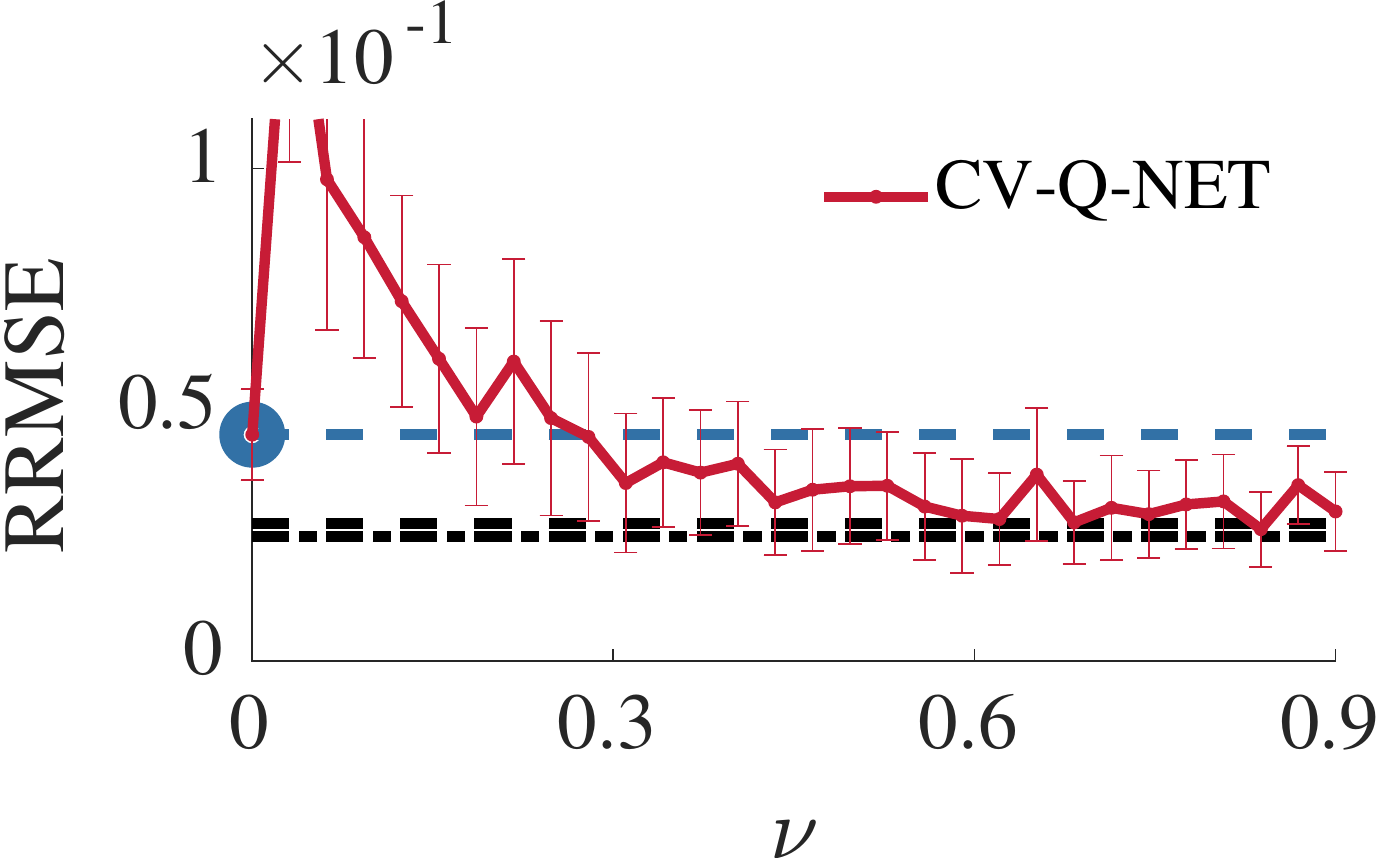}&
		\includegraphics[width=.195\linewidth]{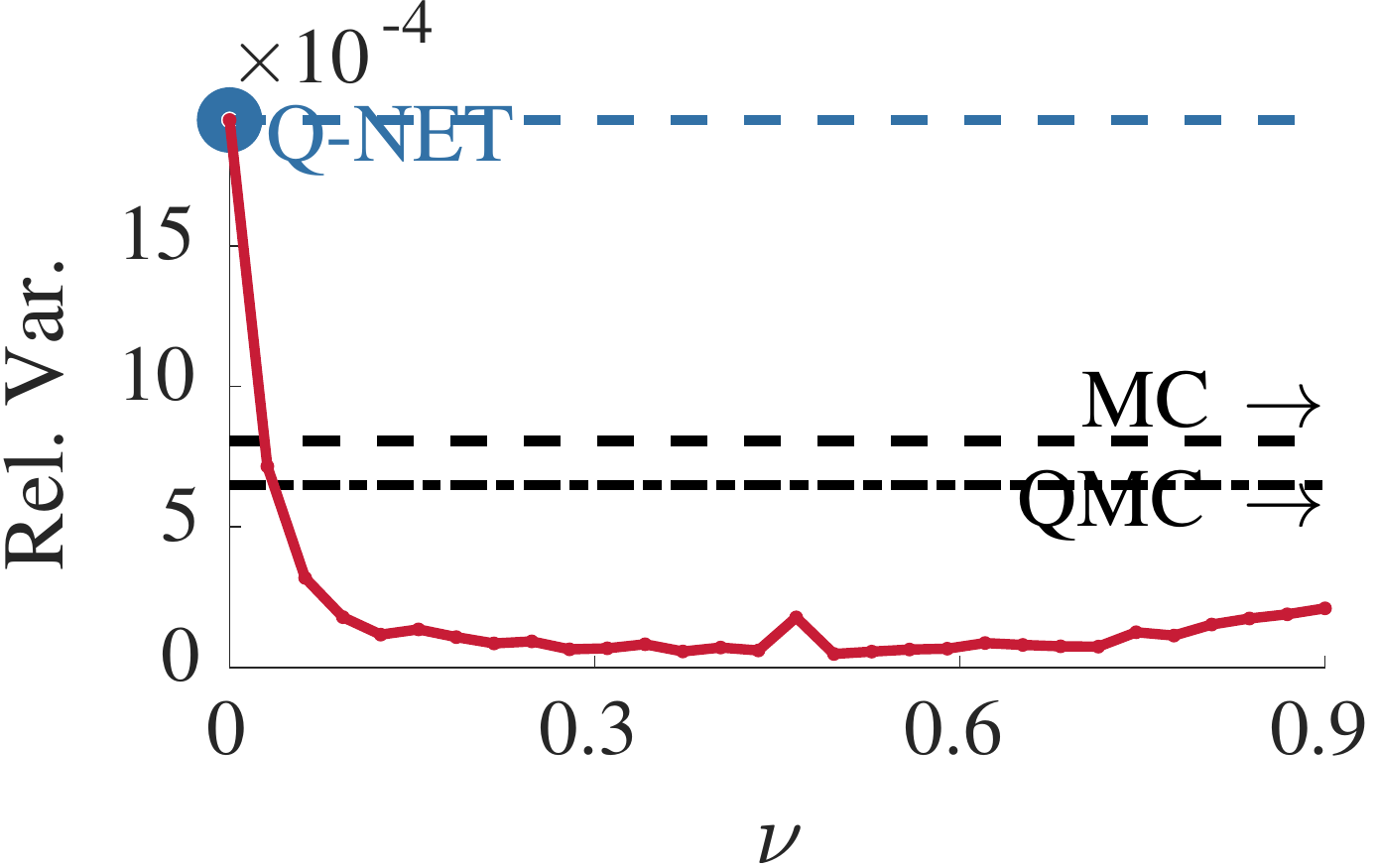}\\
		\multicolumn{2}{c}{(a) 4D HR} &
		\multicolumn{2}{c}{(b) 8D HR}
	\end{tabular}
}{The proxy can be used as a control variate (CV) by using a fraction ($\nu$) of the samples to integrate the difference $\f-\fw$ using MC or QMC. The plots show relative error and variance of the CV estimator (red) compared to Q-NET ($\nu=0$) and MC and QMC ($\nu=1$) estimators. They reveal that CV-Q-NET is effective at reducing variance (lower than MC and QMC) for a wide range of values of $\nu$ in 4D (a) as well as in 8D (b). }{cverrvar}

\hdg{Marginalization}
The projection operator uses the same Q-NET implementation as tested above but on a subset of dimensions.  \figref{marginals} shows a qualitative assessment of integrating a 3D Gaussian mixture with $35$ components using $k=35$ neurons and $N=2048$ samples. The figure omits subscripts on $\mu$ for brevity.  The point-sampled functions $\mu^1$ and $\mu^2$ are plotted for different permutations of the components of \x. The marginals obtained using Q-NETs match references for all 2D and 1D projections. We use marginalization in our example application for Bayesian inverse rendering in~\secref{bir}.

\hdg{Sub-integrals} We trained \fw\ using $k=180$ neurons and $N$ samples of the zone plate function $f(\x) = \left(1+\cos{(220 \; \x^\intercal \x)}\right)/2$ in $[0,1]\times[0,1]$ and estimated integrals over subdomains of different sizes. \figref{subint}  visualizes \f\ (left), the subdomain sizes (white boxes) and plots of RRMSE (right) as $N$ is increased up to $30K$. The plot shows the mean RRMSE along with the standard deviations (shaded region) over $100$ randomly shifted square subdomains with sides $1/3$, $1/8$ and $1/20$. Average error is larger for smaller subdomains, as expected, since the number of expected samples representing the function within the subdomain drops quadratically with respect to the side. That is, for this function, integrals over arbitrary sub-domains as small as $1/20\times1/20$ can be obtained with $10\%$ error if \fw\ was trained using a $170\times170$ grid over the unit square. Error curves for integration over the entire domain (`$1$' in the legend) using Q-NET (solid) and QMC (dotted) are also shown.

%% file: cga.tex
\section{Results II: Sample applications} \label{sec:apps}

We demonstrate the utility of the proposed proxy and its integration via Q-NETs within a few computer graphics contexts. Our aim is to highlight the versatility of the proxy and its potential to inspire future work, rather than to claim improvement over state of the art in a specific application. 

\myfigfull{!ht}{
\begin{tabular}{@{}b{.2\linewidth}@{\hspace{3ex}}b{.36\linewidth}@{\hspace{3ex}}b{.36\linewidth}@{}}
    \raisebox{.04\height} {\includegraphics[width=\linewidth]{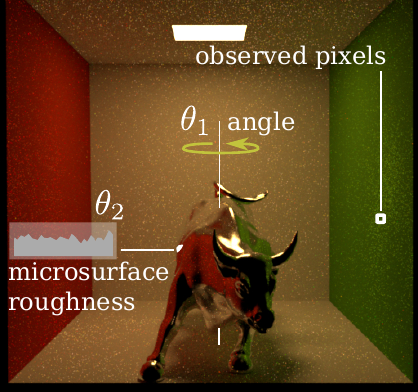}}&
    \includegraphics[width=\linewidth]{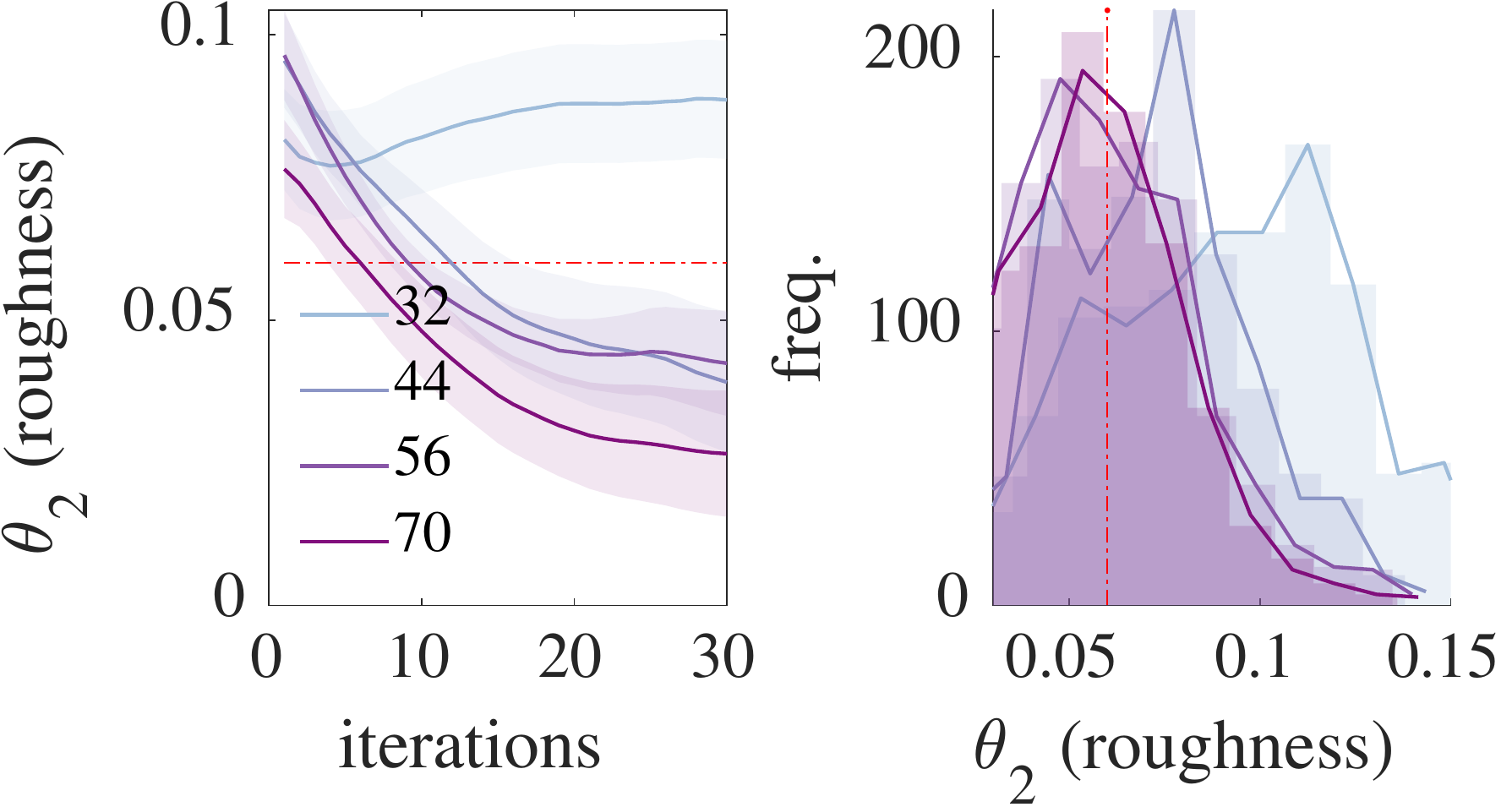}&
    \includegraphics[width=\linewidth]{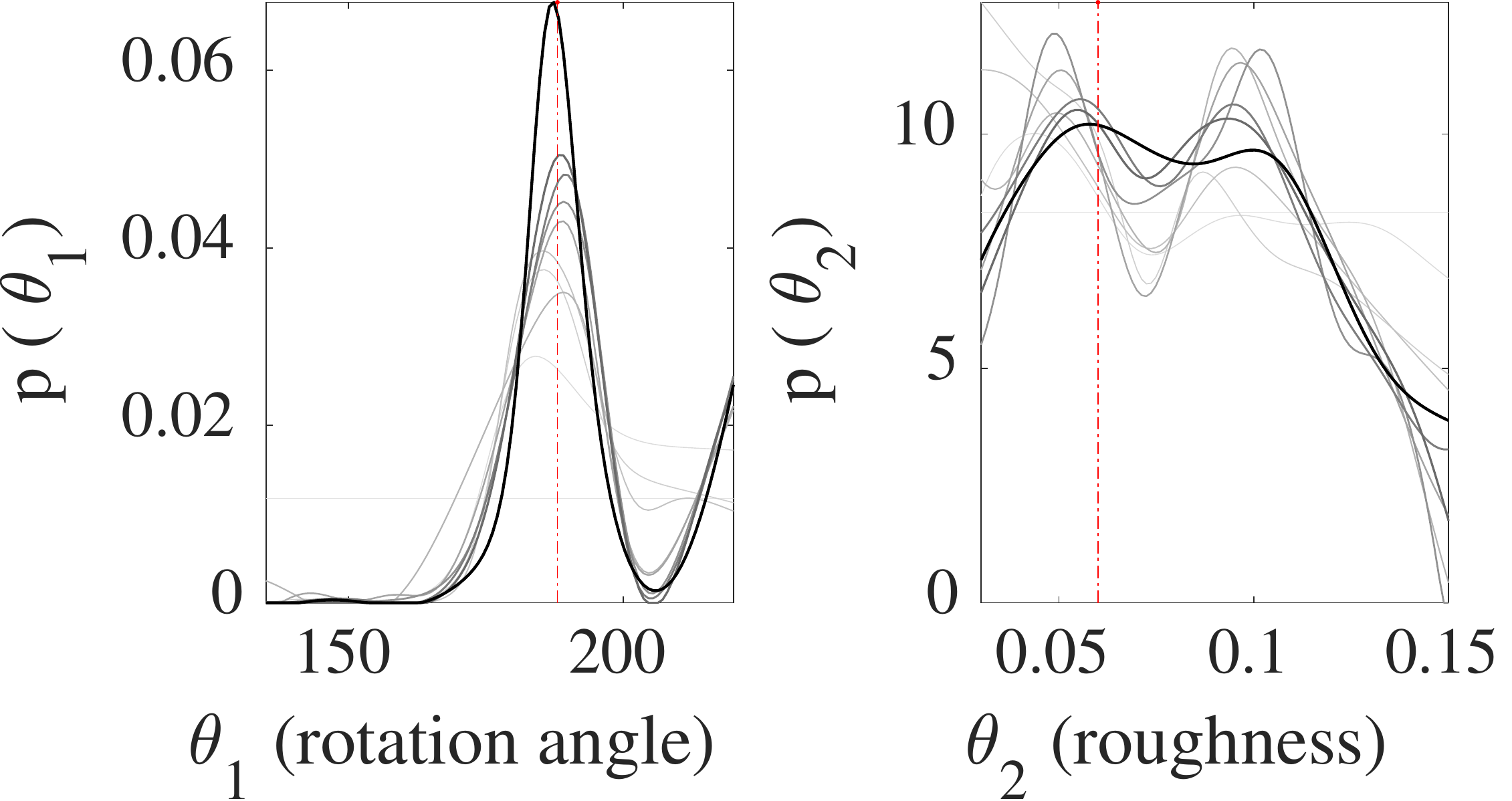}    \\
    \multicolumn{1}{c}{(a) scene and parameters} & 
    \multicolumn{1}{c}{(b) Point estimates for $\theta_2$ using Mitsuba 2} & 
    \multicolumn{1}{c}{(c) Bayesian inference using Q-NET}
\end{tabular}
}{(a) Given observed radiance (patch on the green wall) modern differentiable renderers like Mitsuba 2~\cite{NimierDavidVicini2019Mitsuba2} are effective at inferring point estimates for differentiable scene parameters such as $\theta_2$. (b) We ran several iterations, using different patch-sizes for observations and plotted these estimates and their frequency polygons (histograms). (c) Our method can be used to infer distributions over parameters which may ($\theta_2$) or may not ($\theta_1$) be differentiable. We achieve this by using precomputed radiance samples from a standard forward renderer to train \fw\ to be the 3D joint distribution $\fw(\T, \Lt)$. Then we perform Bayesian inference starting with a uniform prior (light grey). Given unseen observations (on the green wall), the iteratively refined posterior distributions over $\theta_1$ and $\theta_2$ are shown with progressively darker greys and reference values are shown with dashed red lines. (Also explained in the video)}{invrender}

\subsection{Bayesian inverse rendering} \label{sec:bir}
We use a proxy to estimate probability distributions over rendering parameters (such as material roughness and rotation transformations) given only a small patch ($6\times 6$) of noisy, rendered pixels. Several recent methods~\cite{MPDR20,BTDDiff2020,NimierDavidVicini2019Mitsuba2,DRT19} focus on inferring the inputs \T\ to a renderer (parameters such as materials, geometry, etc.) that result in an observed (target) radiance distribution \Lt. This is achieved typically using a differentiable rendering pipeline which allows iterative back-propagation of gradients with respect to \T\ towards optimizing \T. Differentiable renderers are able to infer point estimates of \T\ without prior knowledge or precomputation by virtue of the modified rendering pipeline. They are often not differentiable with repect to subsets of parameters.~e.g.~rotation transformations in Mitsuba 2 ($\theta_1$ in~\figref{invrender}a). Bayesian approaches model the underlying distributions in observed data and are popular for solving inverse problems~\cite{stuart_2010,BInv2020}.

We use precomputation (using a standard forward renderer) to infer the posterior distribution over render parameters given observed radiance. 
\figref{invrender}.a shows an example scene where two parameters of the bull's model were varied: the angle of rotation ($\theta_1$) around the vertical axis and its material roughness parameter ($\theta_2$). 
First, we render images using random vectors of parameters $\T_i$ and record the radiance $\Lt_i$. Then, we train a neural proxy to learn $\fw(\T, \Lt)$ using $(\T_i, \Lt_i)$ and using $k=100$ neurons. We normalize the proxy, using a Q-NET, to estimate the joint probability distribution of parameters and radiance: $p(\T, \Lt)$.
Given observed radiance values \Lo\ and an initial uniform prior $p(\T)$, we iteratively perform Bayesian inference to obtain the posterior $p(\T | \Lo) = p(\Lo | \T) p(\T) / p(\Lt)$. Initially, the numerator $p(\Lo | \T) p(\T)$ is a slice of the proxy at the \T\ where the posterior is evaluated. The evidence $p(\Lt)=\int p(\Lo | \T) p(\T) \mathrm{d}\T$ is a marginal of $p(\T, \Lt)$. For the next iteration, we use the inferred posterior as $p(\T)$, retrain \fw\ to represent $p(\T, \Lt) p(\T)$ and repeat the inference with new observations. The updated posterior marginals over $\theta_1$ and $\theta_2$ are evaluated on (1D) grids and plotted in~\figref{invrender}.c.

Given  observations \Lo\ on the green wall we used a state of the art inverse renderer~\cite{NimierDavidVicini2019Mitsuba2} to obtain estimates for $\theta_2$  (it is not differentiable wrt $\theta_1$).
The plots in \figref{invrender}.b (left) show point estimates vs iterations for different sizes of observed patches. Their histograms hint at the underlying distributions. \figref{invrender}.c plots iterative distributions (progressively darker grey)   using our Bayesian inference over each parameter. Dashed red lines show the reference values used to generate the observed patches \Lo. We trained \fw\ using 100 simulations of a $6\times 6$ crop on the green wall with 100 different $\T_i$. The training time with $50$ and $150$ neurons is $6$ ($8$) and $10$ ($54$) seconds with (without) GPU computation.
After 20 iterations, the modes of the inferred (black) distributions match the reference. Also, the results suggest a low confidence in the inferred $\theta_2$, which is non-trivial to obtain robustly from point estimates. Although our inference scheme does not require a differentiable renderer, it relies precomputation for learning. An animated summary is presented in the accompanying video.


\subsection{Modeling with neural noise}
Sigmoidal approximator networks with random \w\ (no training) are useful generators of noise. They can be  evaluated easily on GPUs, trained to resemble examples, sliced and transformed using intuitive parameter settings. Their integral enables exact filtering, normalization and marginalization. Using Q-NETs, the noise can be sampled with zero evaluations of the noise function.
 
Various classes of procedural noise are usedful in modeling virtual worlds.  Perlin noise ~\shortcite{HTexPerlin89}, a type of lattice-gradient noise, is the de-facto choice due to its visual appeal, easy and efficient implementation and extensibility. Its computational cost is  $O(d2^d)$ per evaluation for noise in $d$ dimensions. Simplex noise~\cite{SimplexN} improves on this, with a cost of $O(d^2)$ per evaluation. Gabor noise is a popular alternative that can be trained from examples~\cite{GaborByEx12} and filtered~\cite{FGab11}. Neural noise can be computed in $O(dk)$ per evaluation if $k$ neurons are used. 

\myfig{!ht}{
\begin{tabular}{@{}cccc@{}}
\includegraphics[width=.243\linewidth]{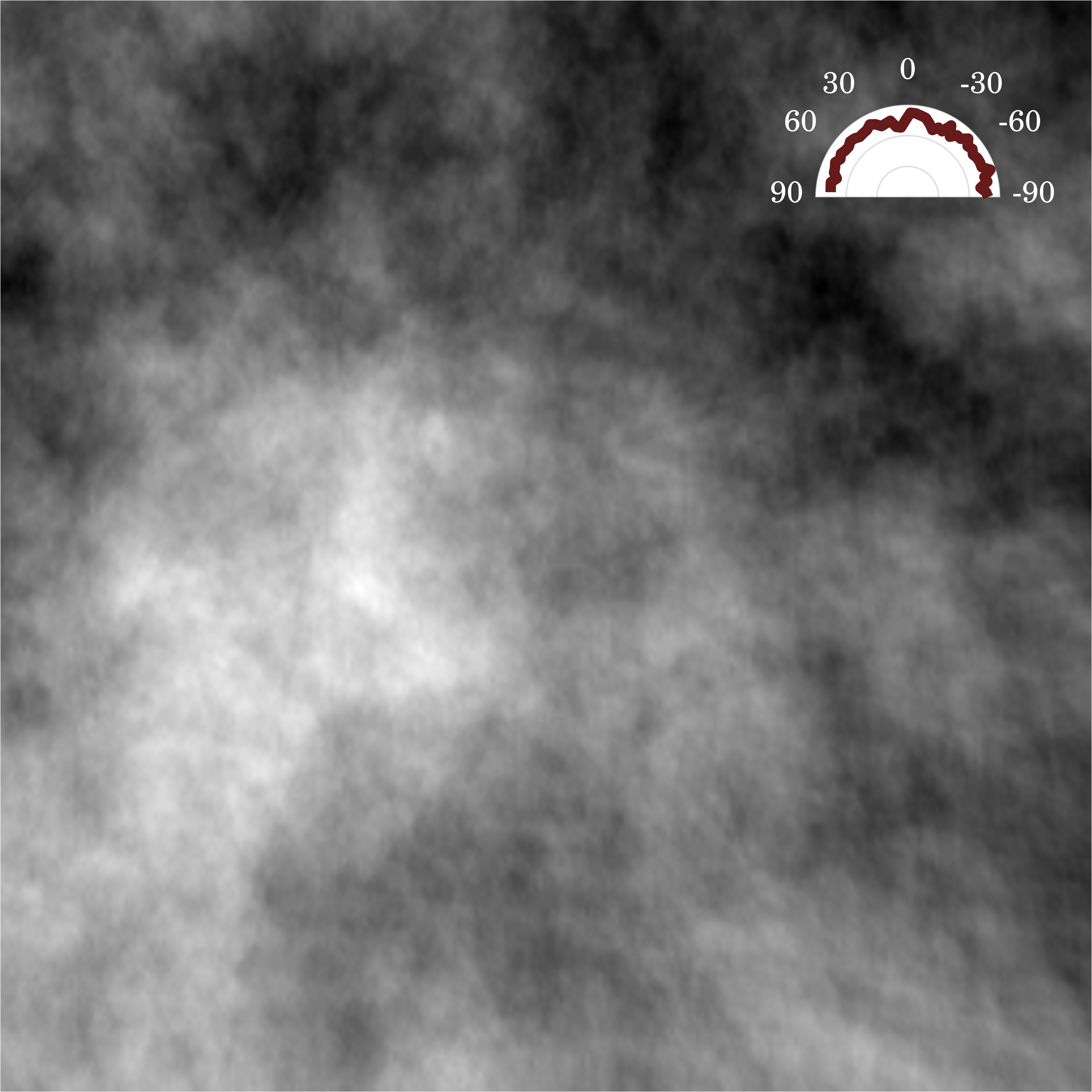}
\includegraphics[width=.243\linewidth]{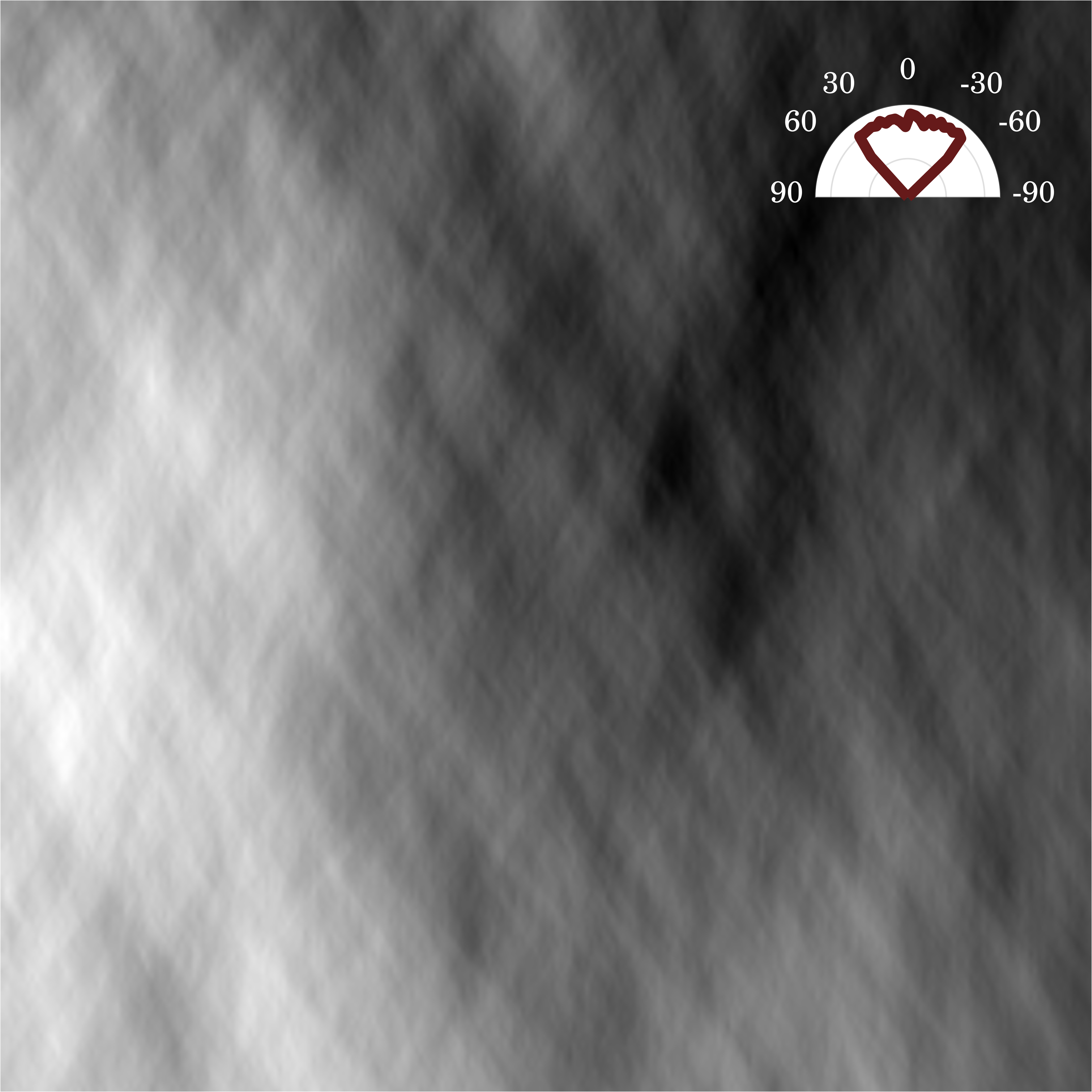}
\includegraphics[width=.243\linewidth]{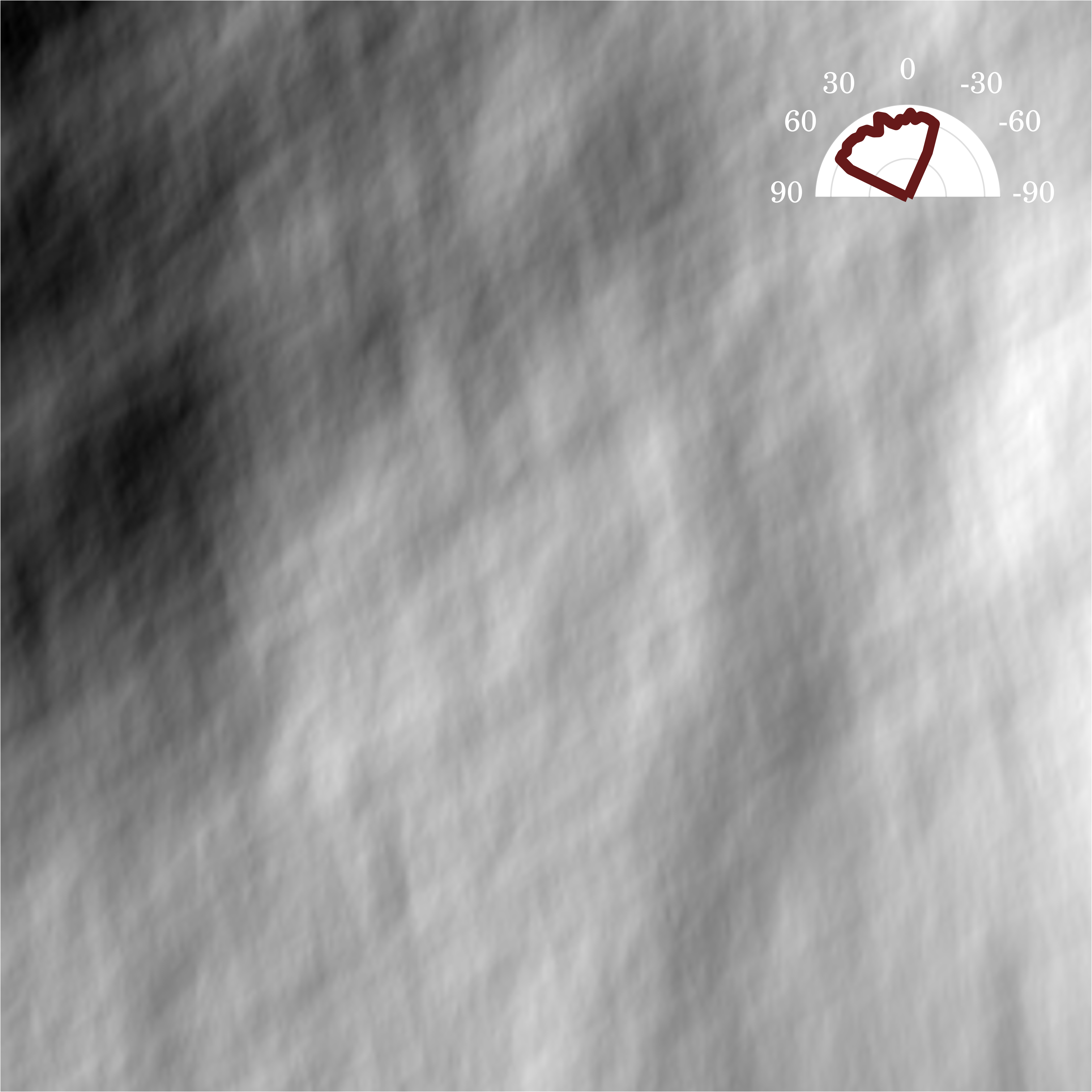}
\includegraphics[width=.243\linewidth]{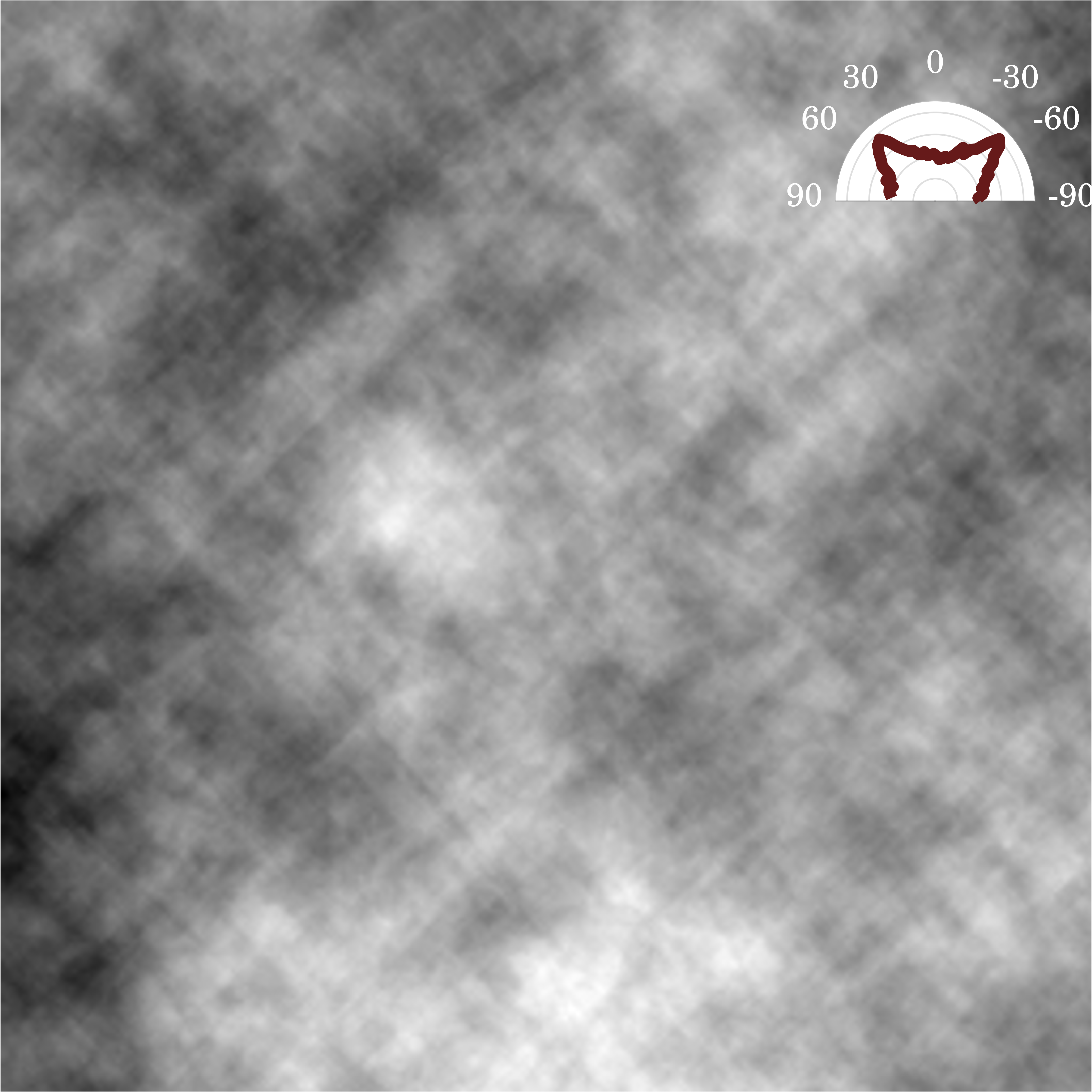} \\
\includegraphics[width=.243\linewidth]{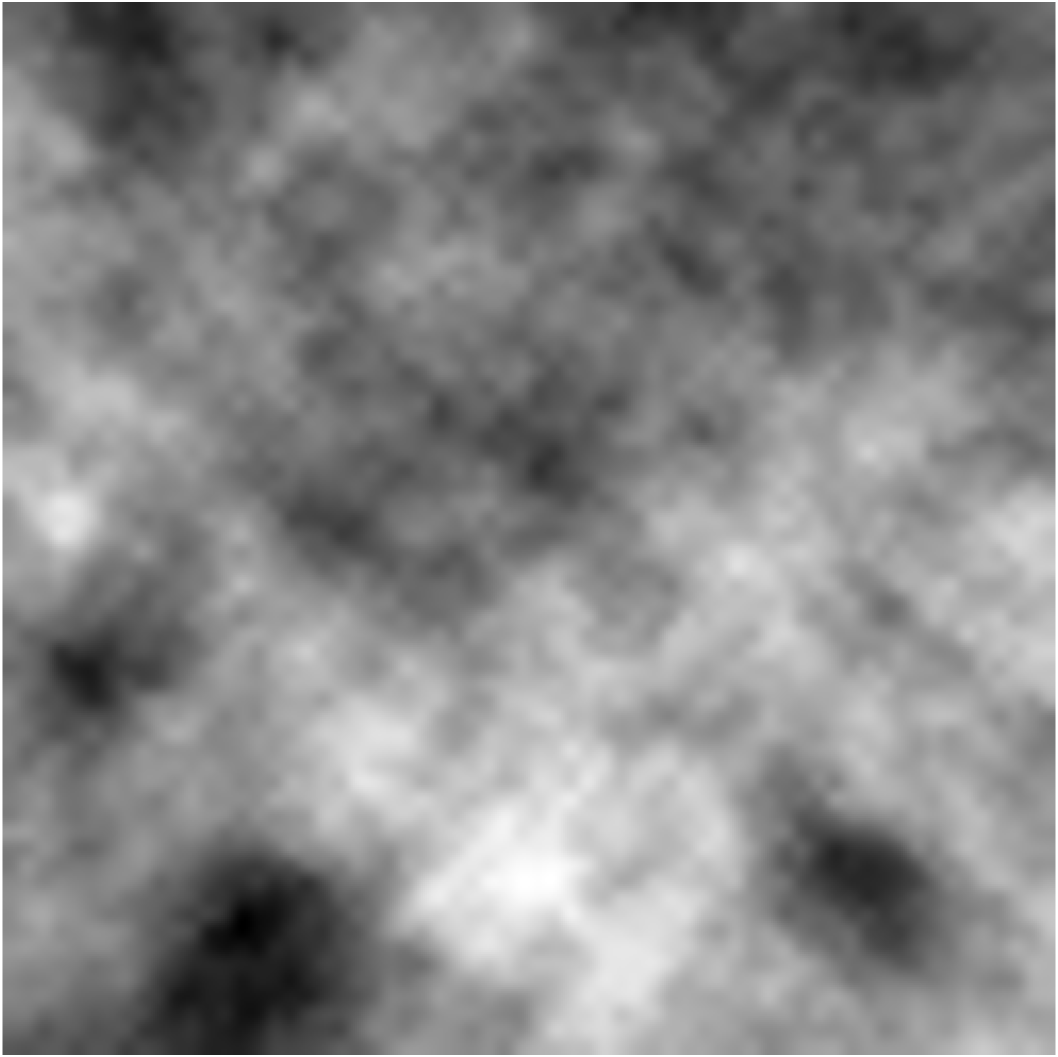}
\includegraphics[width=.243\linewidth]{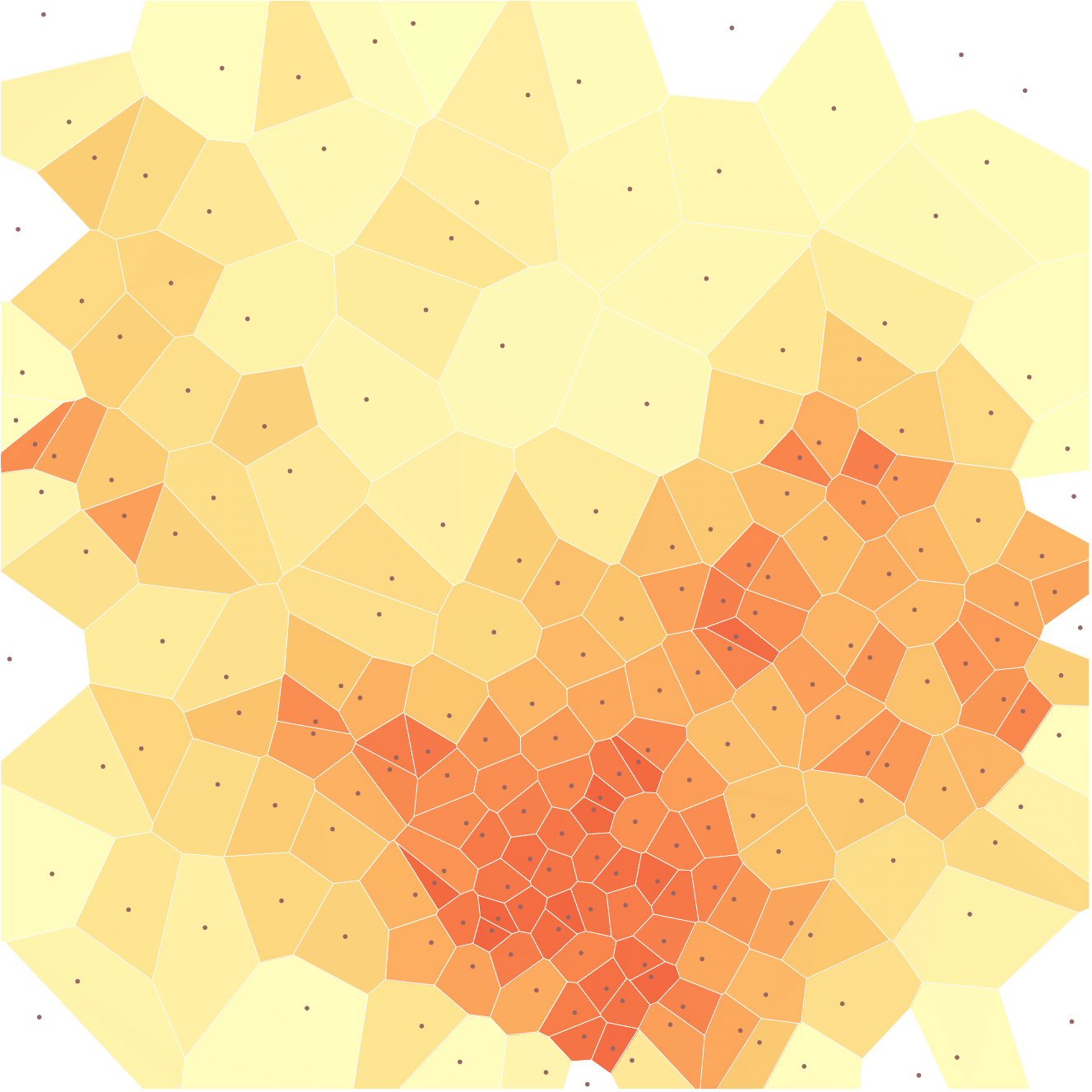}
\includegraphics[width=.243\linewidth]{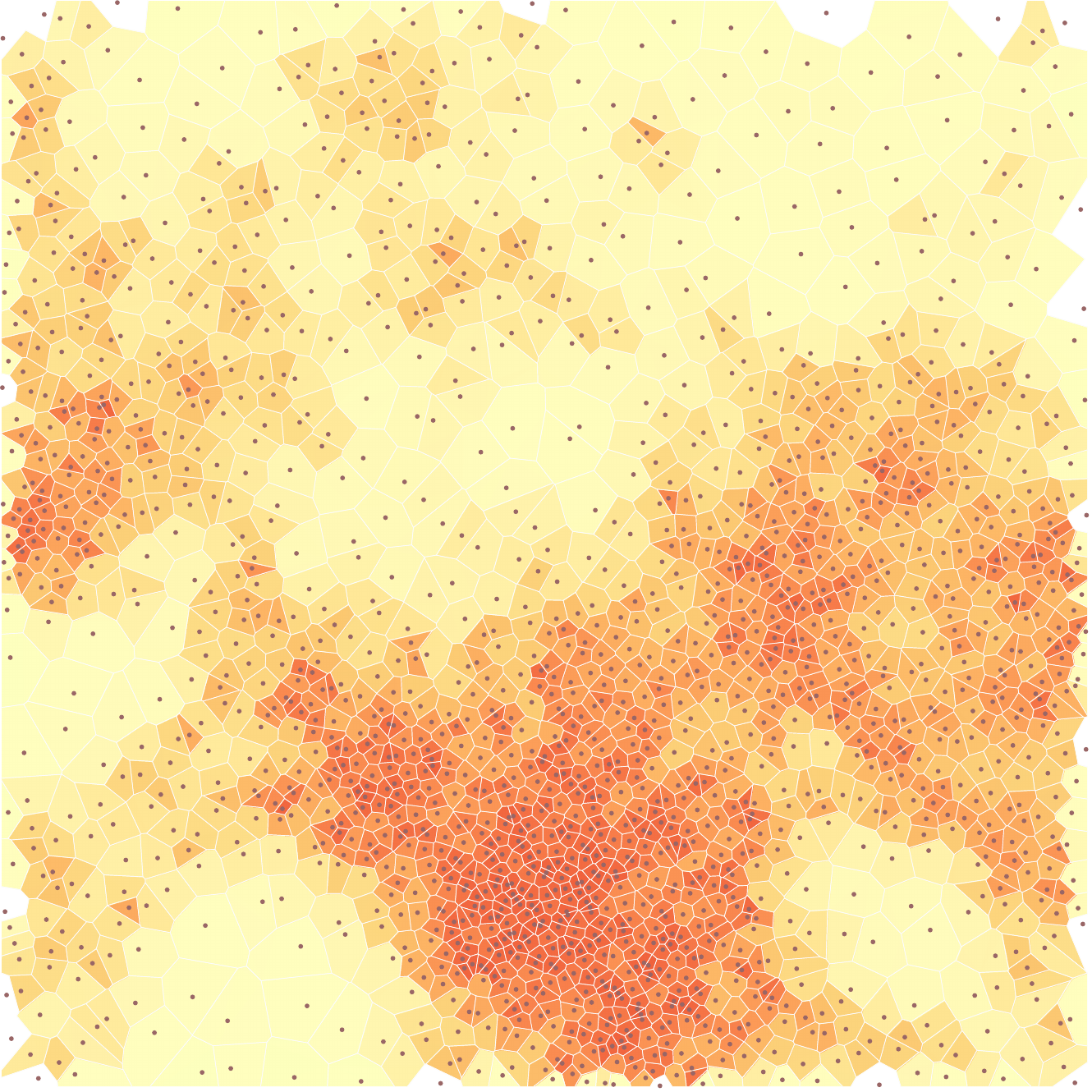}
\includegraphics[width=.243\linewidth]{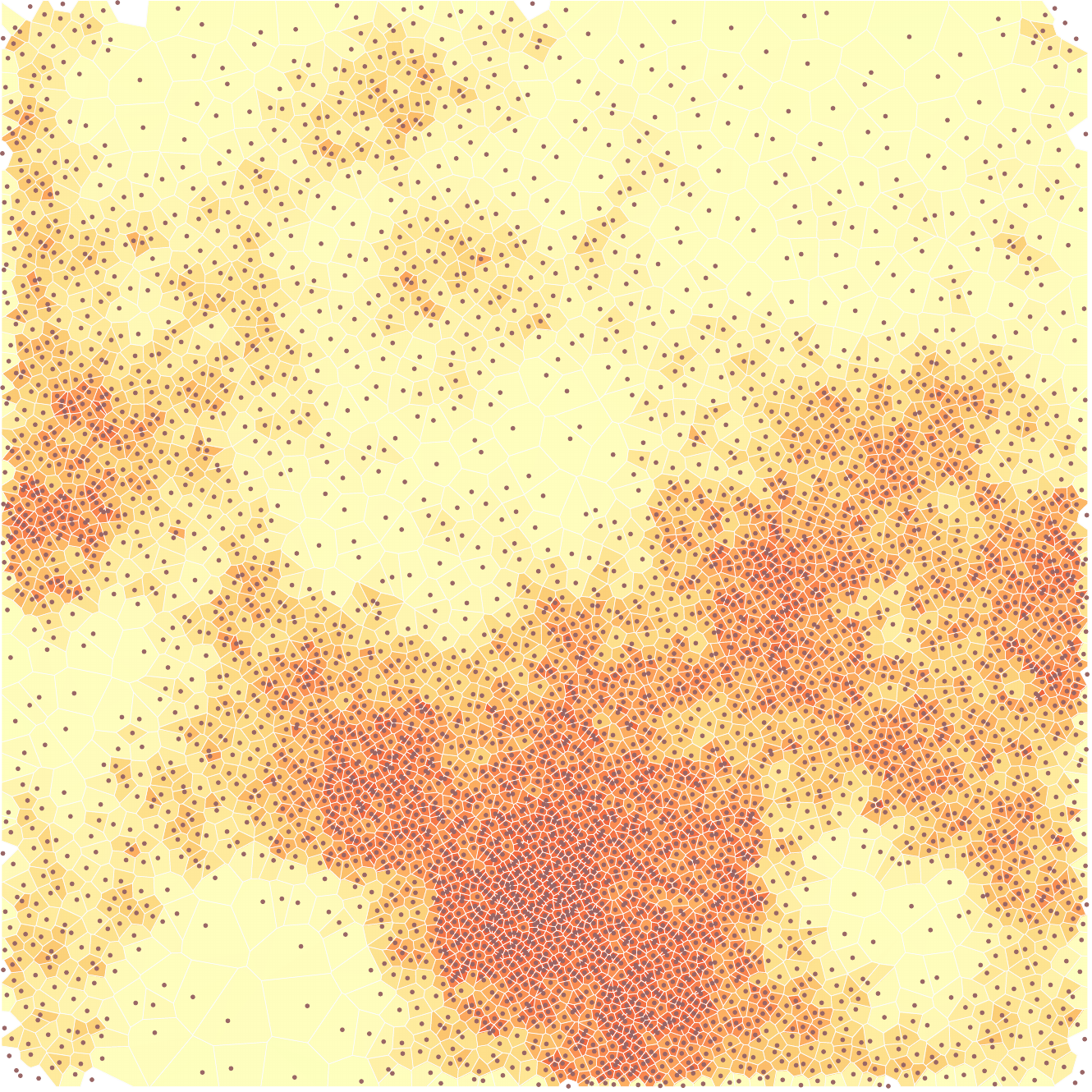}
\end{tabular} 
}{\emph{Top row:} Examples of noise with target anisotropies (insets). \emph{Bottom row:} input 2D noise and samples generated (180, 1500 and 4500 samples).}{noise}

Just as the location $\x_c$ of the middle of the step (where $\sigma(\x_c)=0.5$) is given by $\x_c=-\ba/w_1$ in 1D, the orientation of the $i^{th}$ 2D sigmoid is given by $\wa^{i,2}/\wa^{i,1}$. The random values in \wa\ can be guided (see video) to produce noise functions with different distributions of sigmoid-orientations as shown in the top row of ~\figref{noise} ($k=25K$). The insets visualize the distribution of orientations. Evaluation speed is about $13$ KHz, $30$ KHz and $500$ KHz for $25K$,  $10K$ and $100$ neurons respectively (averaged over 1 million evaluations). 

The bottom row in \figref{noise} visualizes a target noise pattern and 180, 1500 and 4500 Halton samples respectively, along with their Voronoi cells. The samples (best viewed by magnifying on a screen) were generated via proportional allocation over gridded tiles with the number of samples in each tile proportional to the integral of the noise within that tile (calculated using a Q-NET). The cells may also be generated via a k-d tree type construction where partitioning values at each step are the medians of the respective functional marginals obtained via a Q-NET. 
 
Neural noise patterns scale well to multiple dimensions. The submitted video shows a scene containing a time-varying participating medium with heterogeneous density (3D + 1D).  We modeled the 3D density in each frame as a slice (in time) of a 4D noise function with $k=10K$ neurons. We rendered images using standard MC path tracing implemented in PBRT~\cite{PBRT3e}. Q-NETs enable analytical integration of optical depth along rays (\secref{optdepth}).

\subsection{Estimating optical depth}\label{sec:optdepth}
We represent a scalar density field using the neural proxy and integrate the density along rays by first performing appropriate transformations (translation to the ray origin and rotation to align one of the dimensions with the ray direction) and slicing. 
The natural logarithm of the ratio of incident to transmitted radiant power through a material, or \textit{optical depth} $\tau$, often needs to be estimated for rendering and volume visualization applications.  The optical depth between two points is usually calculated via integration of $\alpha(s)$, the attenuation coefficient at a distance of $s$ from the first point, over the line segment between the points. The integrals are usually estimated via sampling or ray-marching for heterogeneous media. 

\myfig{!ht}{
\begin{tabular}{@{}c@{\hspace{.3ex}}c@{\hspace{1 ex}}c@{}} 
    \includegraphics[width=.215\linewidth]{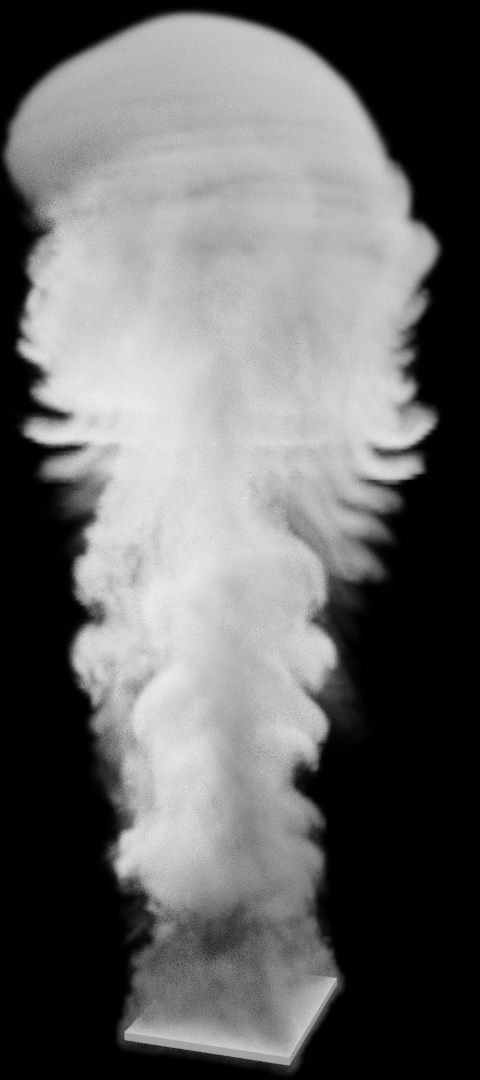}&
    \includegraphics[trim=0 10 0 0, clip, width=.233\linewidth]{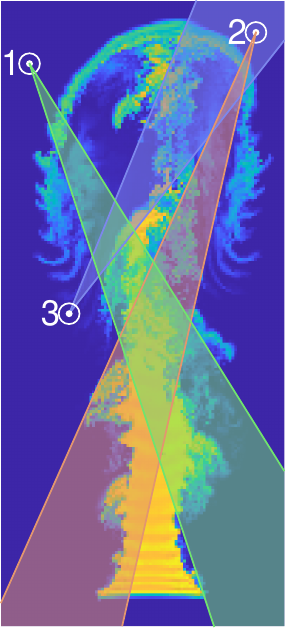}&
    \includegraphics[width=.51\linewidth]{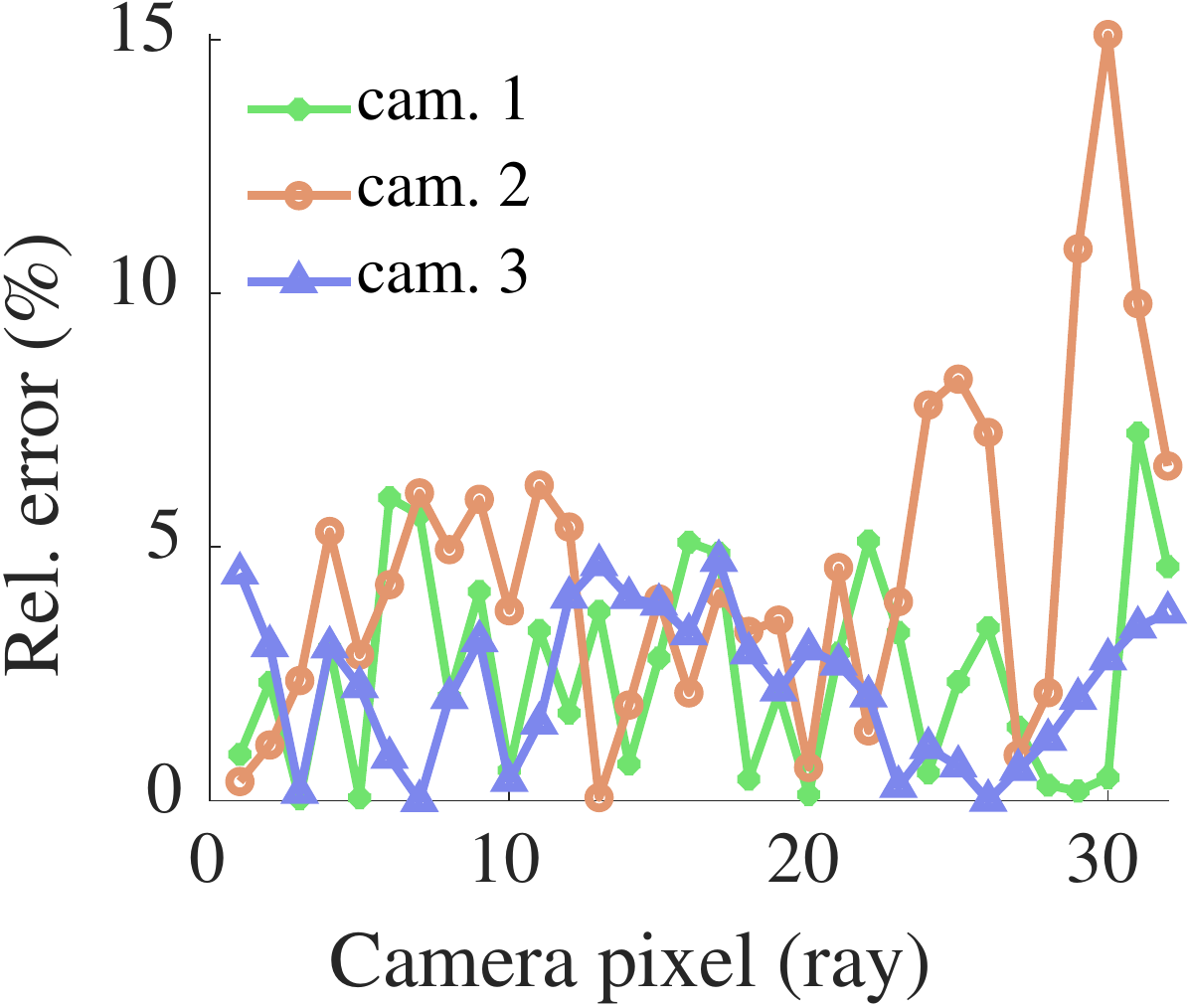}
\end{tabular} 
}{We fit a proxy to density data from a smoke simulation in Blender and calculate optical depth along rays using a Q-NET. A frame from the simulation is shown rendered using Blender Cycles (left). For the central slice, we use a value proportional to the density as the attenuation coefficient and integrate it along $32$ rays each for three virtual camera poses (middle). The plot (right) shows percentage relative error for the rays. }{smoke}

We performed a smoke simulation using Blender, exported the $101\times 100\times 223$ spatially-varying density to a VDB file and trained \fw\ within the support volume (simulation domain). Given a ray specified by its origin and direction, we derive the transformation that aligns the $X$ axis of \fw\ with the ray. Then, we slice \fw\ with $y=0, z=0$ and integrate along the remaining dimension to obtain the optical depth along the ray within the volume. 
\figref{smoke}.a visualizes a frame of the smoke simulation rendered using Blender cycles. \figref{smoke}.b visualizes the central vertical slice of the simulation along with three virtual 2D cameras. For each of the $32$ pixels per camera, we estimated optical depth (until the ray leaves the volume) and plotted the relative errors (as percentages) in \figref{smoke}.c. The plots are representative of our experiments: $90\%$ of the estimates were below $5\%$ relative error and about $98\%$ of the estimates are below $10\%$.

\subsection{Visualizing flux through rectangular volumes}
The flux $J_s$ of a vector field $\Fv(\x)$ through a cuboidal volume (voxel) of side $s$ is usually calculated as a surface integral of \Fv\ over the surface of the voxel. According to the divergence theorem,
\vspace{-.5em}
\begin{equation}    
 \label{eq:flux}
    J_s = \DefInt{V_s}{}{\triangledown.\Fv(\x)}{\x}
    \vspace{-.5em}
\end{equation}
where $V_s$ is the volume of the voxel and $\triangledown.\Fv(\x) \equiv \sum \partial \Fv/\partial \x^i $ is the divergence of \Fv. Thus, if \fw\ is trained to represent divergence then an approximation to the above integral can be calculated in closed-form for cuboidal neighborhoods. \figref{vfviz} shows this calculation applied to a medical dataset of deformation observed in a human lung via 4D CT scans~\cite{vandemeulebroucke2007popi}. The dataset provides a 3D deformation field during breathing, which we use as \Fv. The divergence of \Fv\ then corresponds to the trace of the local strain tensor which is called \emph{dilatation} (or dilation). The streamtube plots in \figref{vfviz}b. and c. visualize local dilatation between two 3D frames (inspiration and expiration) along streamlines in \Fv.  In \figref{vfviz}d. we color streamtubes by the local flux within $5\times5\times5$-voxel neighborhoods. In this context, flux corresponds to average expansion (blue) or compression (red) of local neighborhoods during expiration. This method could also be useful to represent, interpolate and visualize flux in Lagrangian fluid simulations. 

\myfig{!ht}{
\begin{tabular}{@{}p{.15\linewidth}m{.8\linewidth}@{}}
    \begin{tabular}{@{}c@{}}
    \includegraphics[width=\linewidth]{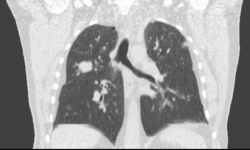} \\
    \includegraphics[width=\linewidth]{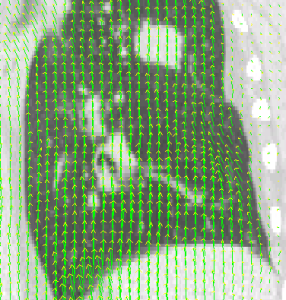}
    \end{tabular} &
    \includegraphics[trim=25 5 55 60 , clip, width=\linewidth]{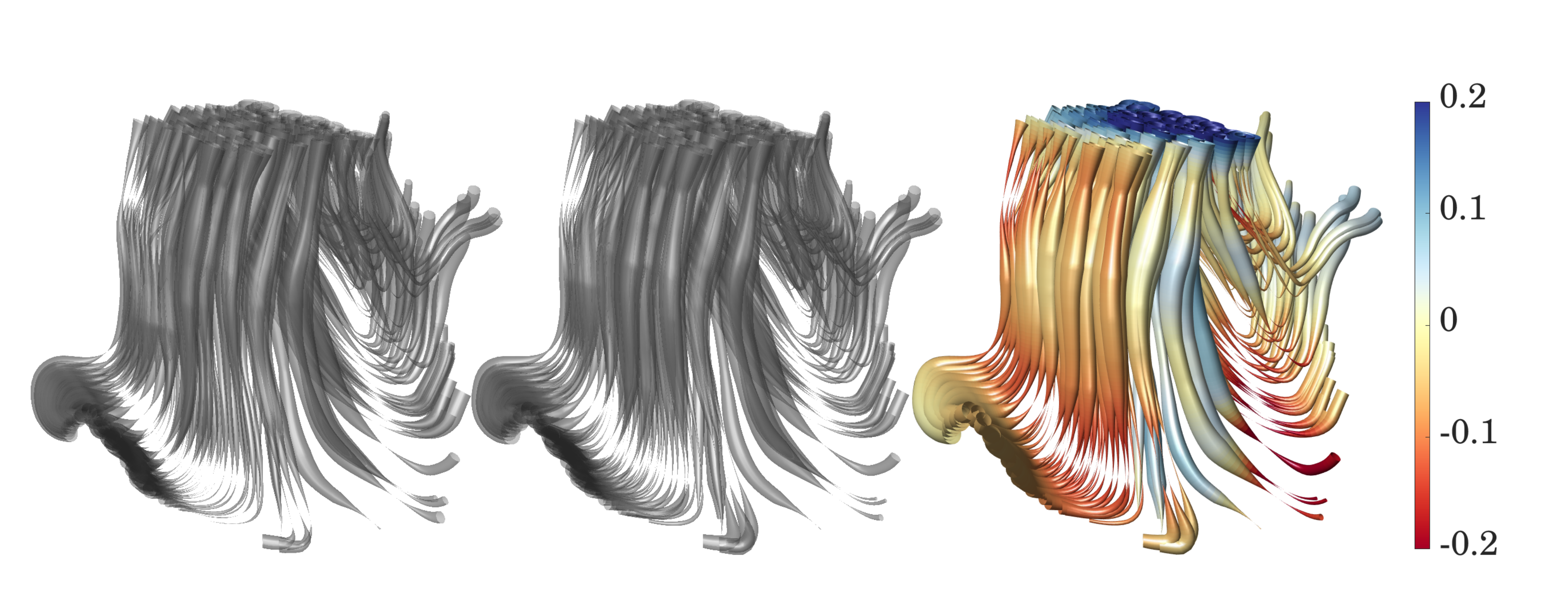} 
\end{tabular} \\    
\begin{tabular}{@{}c@{\hspace{2.2em}}c@{\hspace{1.4em}}c@{\hspace{.9em}}c@{}}
 (a) data & (b) dil. (ref.) & (c) dil. (ours)  & (d) dil. + flux (ours)  
\end{tabular} 
}{(a) Images showing the deformation field in a human lung during respiration~\cite{vandemeulebroucke2007popi}. (b) Streamtubes for deformation between two frames of the dataset. The thickness of tubes corresponds to local dilatation. (c) Approximation dilatation obtained using a neural proxy. (d) Visualizing local flux (color) calculated by integrating the proxy.}{vfviz}

\subsection{Representing signed distance fields}
The derivative of \fw\ wrt any $\x^i$, like the integral, may also be obtained analytically as the weighted sum of derivatives of sigmoids: $\sigma(.)(1- \sigma(.))$. This is exploited in training sigmoidal approximators, but can also be useful to estimate analytical gradients of the proxy.~e.g.~to approximate signed distance fields (SDF).  The utility of the proxy is twofold here: 1) it can be trained using an adaptively sampled SDF; and 2) the approximate SDF and its gradient can be evaluated at arbitrary points in the domain. 

An SDF is a scalar field whose absolute value at any point \x\ can be interpreted as the distance of \x\ to some point $\mathbf{s}$ on surface \Ss. The field is negative if the vector from \x\ to \Ss\ is along the direction of the surface normal at $\mathbf{s}$ and positive otherwise. In addition to SDFs serving as useful implicit representations for modeling~\cite{Blinn1982,SDFSurvey06}, they have also gained popularity as a representation for dynamics simulations~\cite{Koschier2017,Bender19}. \figref{sdfgrad} (top row) shows a bunny and a chair model and ($16^3$) training points where the sampled SDF was used to build a neural proxy. Histograms alongside the models depict reconstruction error of the SDFs tested at ($32^3$) other locations in the volume. About 80\% of the tested points exhibit less than 10\% relative error. 

The middle row of \figref{sdfgrad} visualizes gradients when \Ss\ is a square (centred) of unit sides. The proxy is trained with $2K$ samples in a tubular neighborhood of $0.1$ around \Ss\ along with $200$ samples near the edge of the $[-1, 1]$ domain. Gradient vectors calculated analytically using the adaptively sampled proxy (red arrows) are compared with densely sampled finite-difference gradients (grey arrows). The `zero-set' of the upsampled SDF from which the finite-difference gradients were calculated is shown as a black curve. The magnified images in \figref{sdfgrad} (bottom row) attest to the fidelity of the analytical gradients.The training time was about 2.5 min. for $16^3$ samples with $k=1500$ but evaluation is efficient once trained ($>4MHz$) over $32^3$ evaluations.
%
State-of-the-art methods that are dedicated to solving these problems with SDFs may also benefit from the proxy. e.g. hp-Adaptive grids~\cite{Koschier2017} operate by fitting local polynomial bases within each adaptively subdivided cell of the volume. A limiting step in their method is the calculation of polynomial coefficients by estimating projection integrals.

\myfig{!ht}{ 
\begin{tabular}{@{}c@{}c@{\hspace{2em}}c@{}c@{}}
    \includegraphics[width=.25\linewidth]{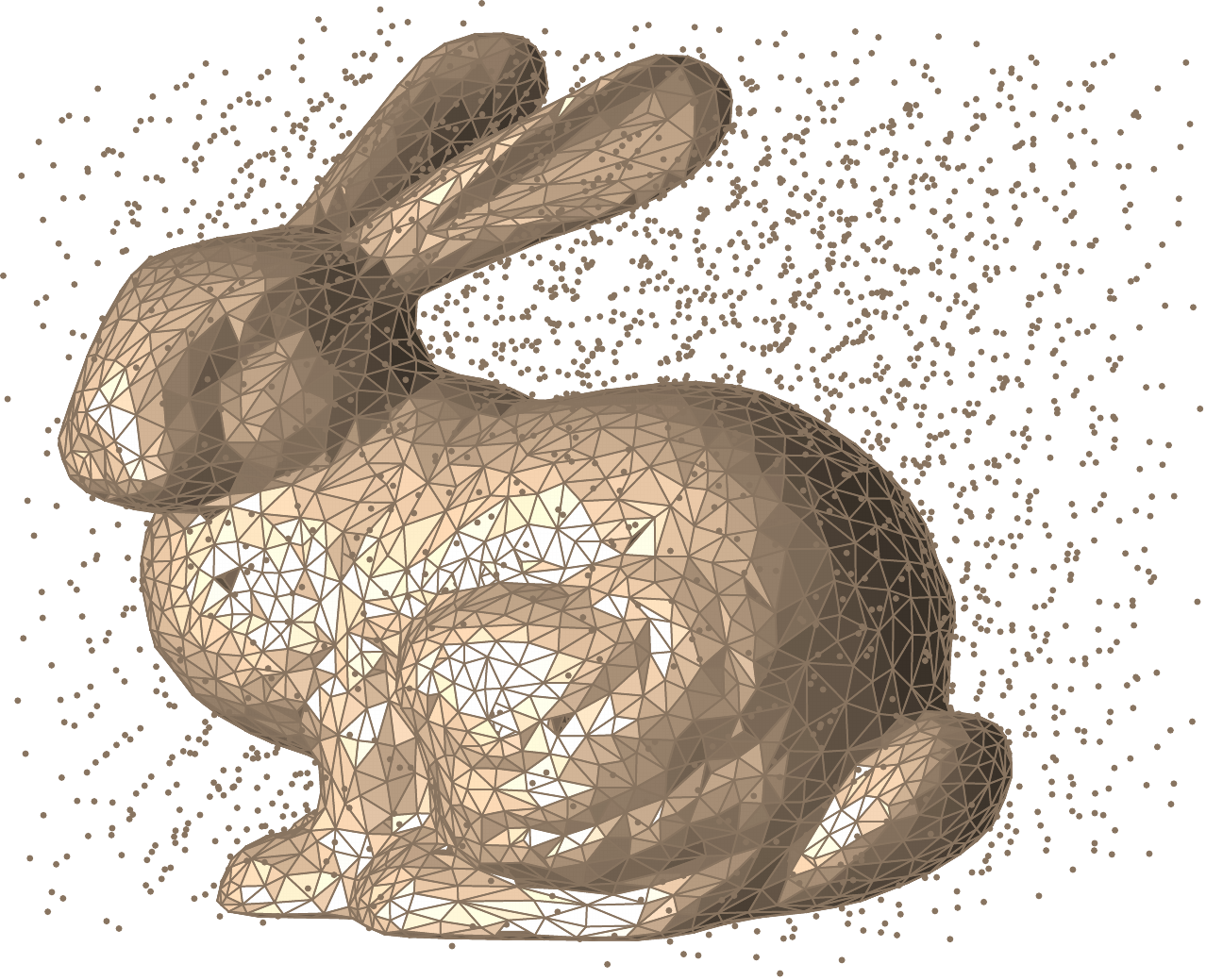}&
    \includegraphics[width=.25\linewidth]{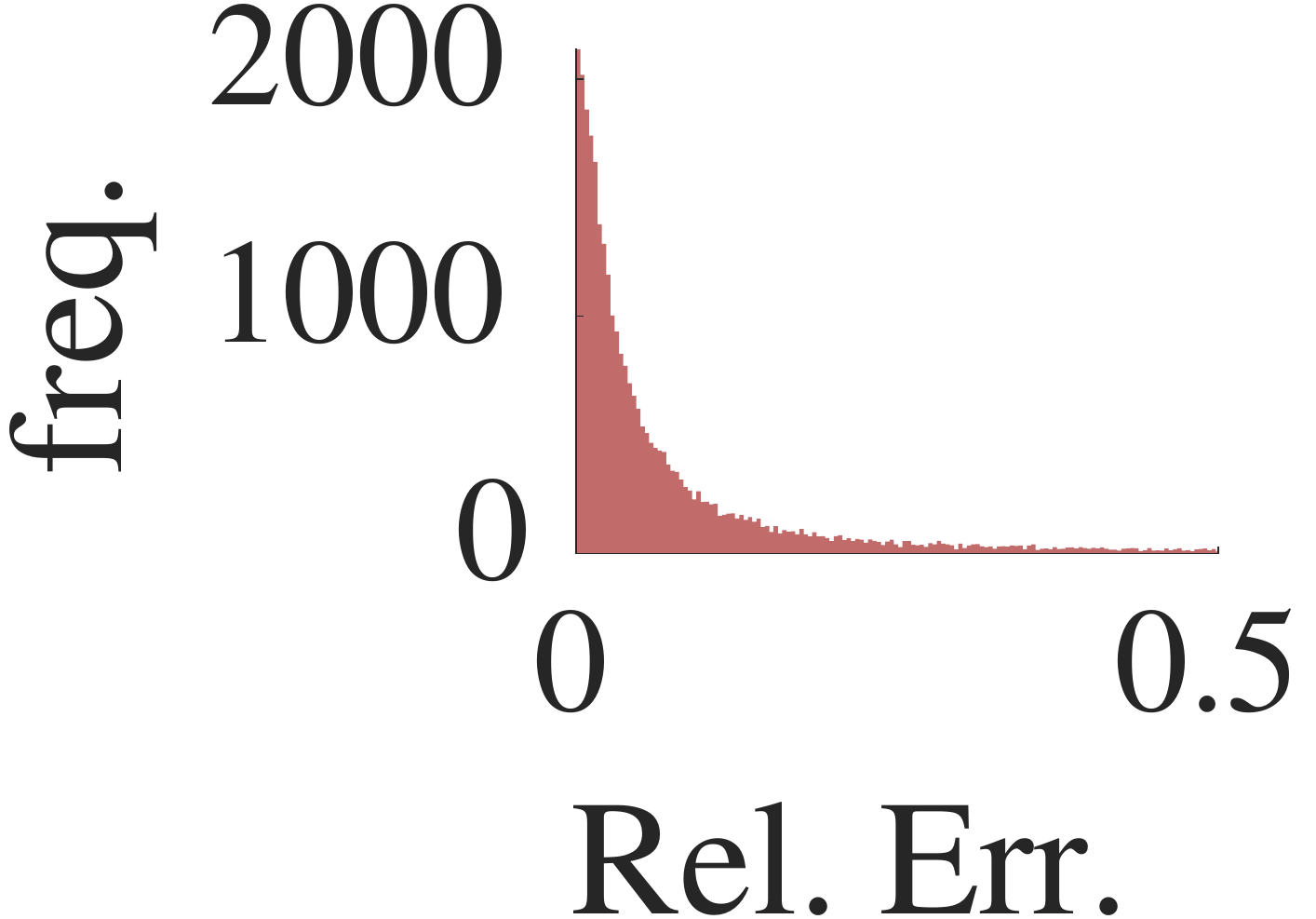}&
    \includegraphics[width=.165\linewidth]{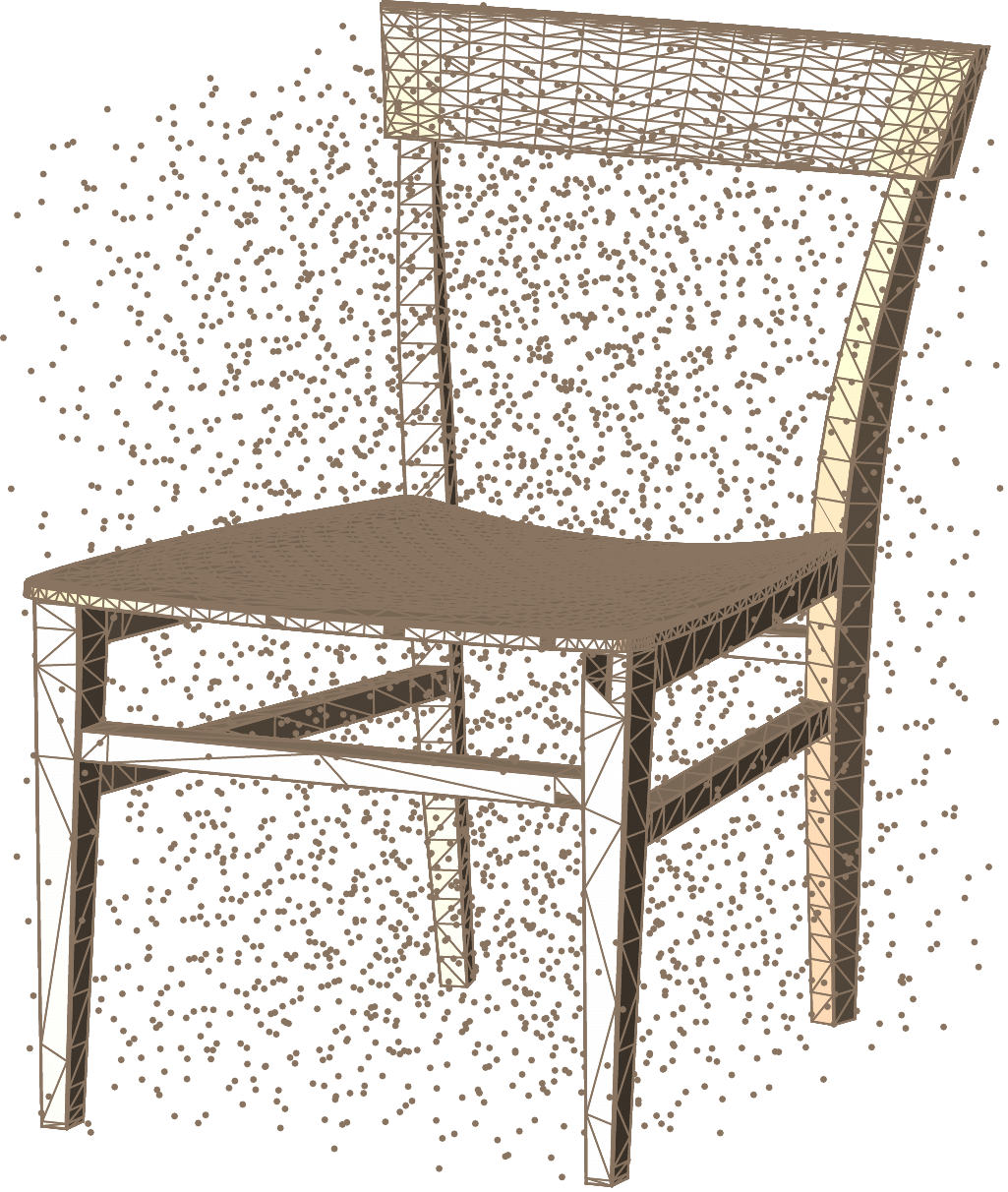}&
    \includegraphics[width=.25\linewidth]{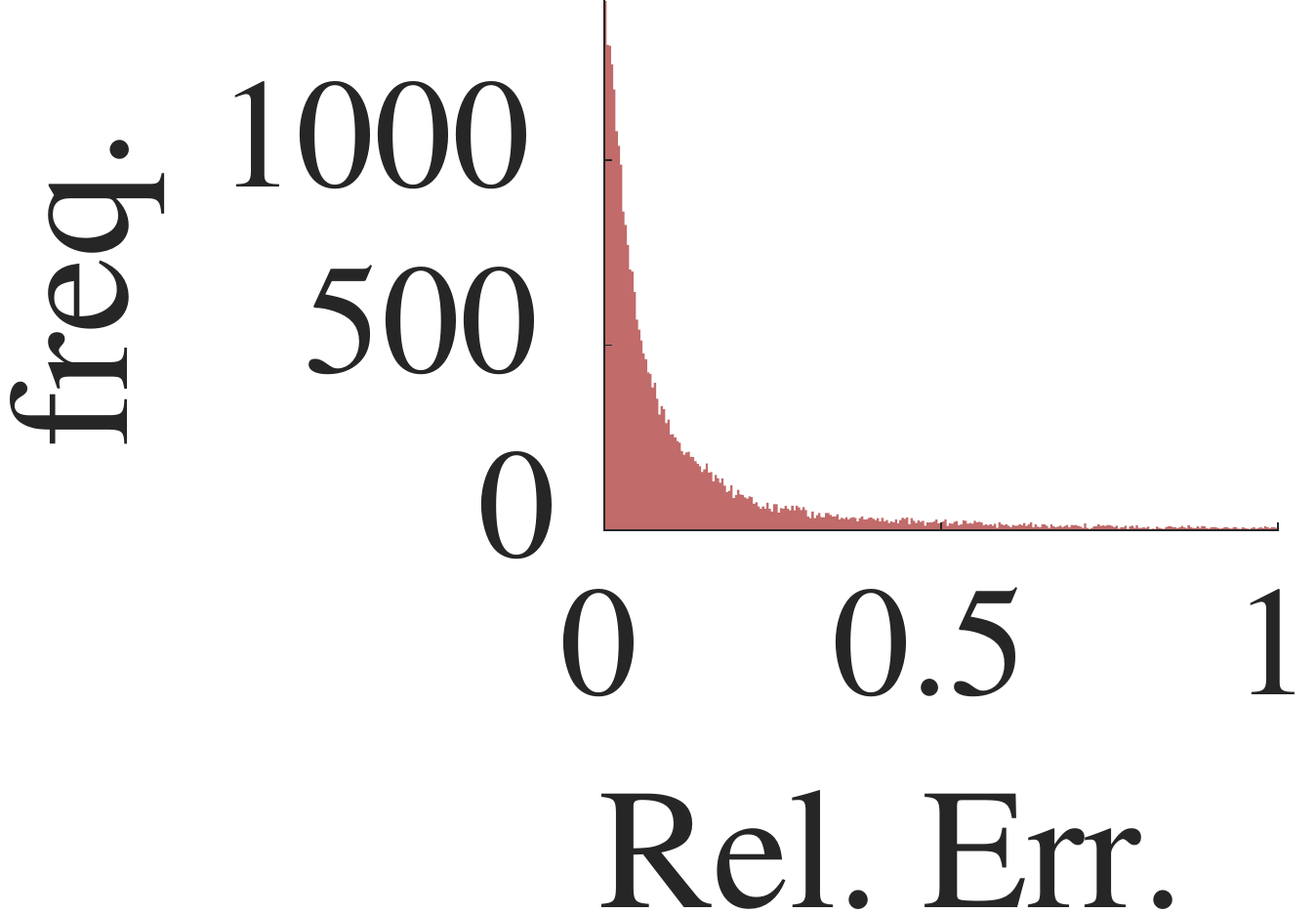} 
    \end{tabular}
\begin{tabular}{@{}ccc@{}}    
    \includegraphics[width=.3\linewidth]{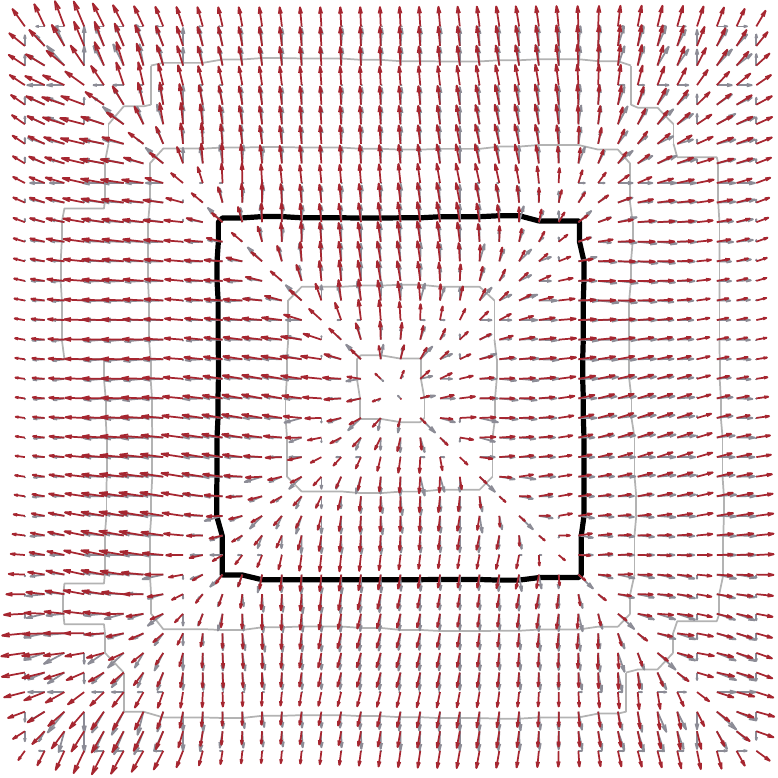}& 
    \includegraphics[width=.3\linewidth]{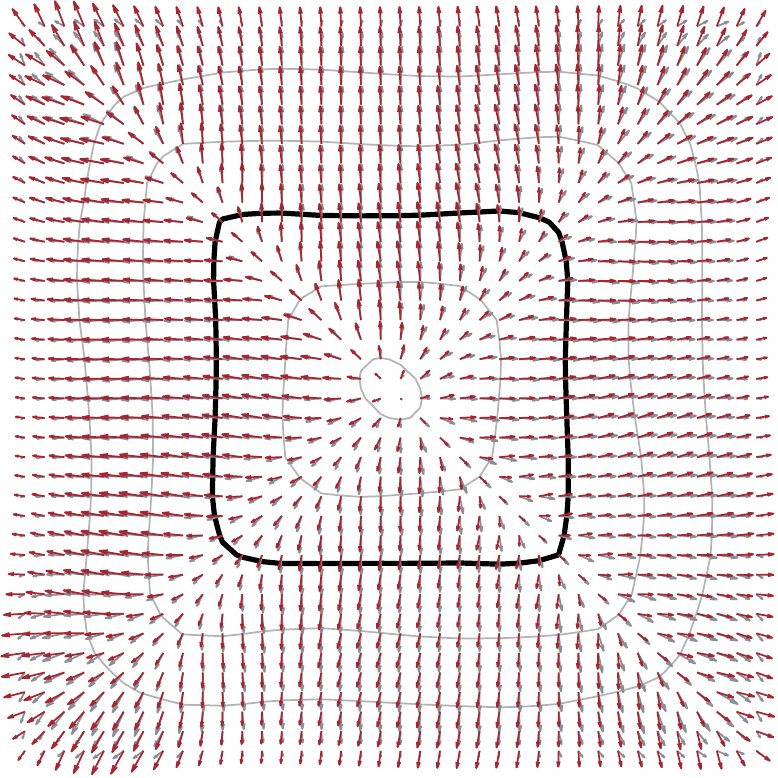}&
    \includegraphics[width=.3\linewidth]{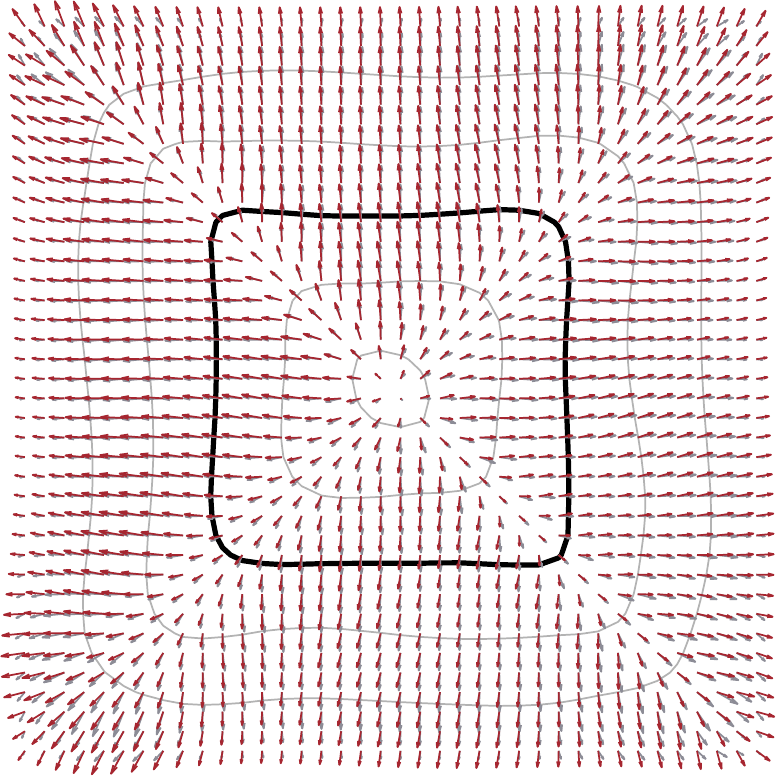} 
    \\
    \begin{tabular}{@{}c@{}c@{}}
    \includegraphics[trim=10 155 165 20, clip, width=.15\linewidth]{BoxSDFGrad_nearest} 
    \includegraphics[trim=125 55 50 120, clip, width=.15\linewidth]{BoxSDFGrad_nearest}& 
    \end{tabular}&
    \begin{tabular}{@{}c@{}c@{}}
    \includegraphics[trim=10 155 165 20, clip, width=.15\linewidth]{BoxSDFGrad_bilinear} 
    \includegraphics[trim=125 55 50 120, clip, width=.15\linewidth]{BoxSDFGrad_bilinear}& 
    \end{tabular}&
    \begin{tabular}{@{}c@{}c@{}}
    \includegraphics[trim=10 155 165 20, clip, width=.15\linewidth]{BoxSDFGrad_bicubic} 
    \includegraphics[trim=125 55 50 120, clip, width=.15\linewidth]{BoxSDFGrad_bicubic}& 
    \end{tabular}
    \\
    (a) ours vs nearest & (b) ours vs bilinear & (c) ours vs bicubic
\end{tabular} \\

}{We fit a proxy with scattered samples from signed distance functions (SDFs) of a bunny (top left), chair (top right) and a square (middle). The square is sampled adaptively (non-uniformly). Error histograms (top row) show that the proxy faciliates a reasonable approximation while enabling scattered interpolation. Gradients may  be calculated analytically (red arrows), which match interpolated gradients well (bottom row).}{sdfgrad}

%% file: discussion.tex
\section{Discussion and future work} \label{sec:discussion}

\hdg{Domain and range transformation}
To generalize the normalized domain of $[-1,1]^d$ and range of $[0,1]$, for example in 1D to $x\in [x_a, x_b]$ and $\fw (x) \in[z_a, z_b]$,  if the normalized function is $\hat{\fw}(\hat{x})$:
$${\fw}(x) = \frac {z_b-z_a} 2 \left( \hat\fw\left(\frac{x_b-x_a} 2(\hat{x}+1) + x_a\right) + 1 \right) + z_a.$$
An estimate $\hat\mu_{d,k}$  in the normalized space can be transformed to an estimate $\mu_{d,k}$ in the true domain via variable substitution as $ \mu_{d,k} = \Omega \;.\; [ (z_b-z_a)  (\hat\mu_{d,k}/2^d +1)/2 + z_a]$, where $\Omega = \prod_{j=1}^d (\x_b^j - \x_a^j) $ is the volume of the domain. We used this in all our empirical tests.

\hdg{Loss functions and training}
We did not notice significant differences in the convergence rate for different loss functions such as \verb+mse, +\verb+mae+ and \verb+cross-entropy+. We found that optimizing using the Levenberg-Marquardt method and Bayesian Regularization perform better than conjugate gradient based methods particularly for discontinuous \f. We used the latter in our experiments because they are suited to training on GPUs.

\hdg{Sampling}
Q-NETs yield a consistent estimator regardless of the sampling distribution provided it is non-zero everywhere that \f\ is non-zero. Low-fidelity reconstruction due to poor sampling could indeed increase variance. Variance is defined across estimates $\mu(\w)$ from samples $\{\x_n\}$ due to stochasticity in the initialization or optimization. The error bars in our plots were generated using a fixed set of samples across repetitions for Q-NETs while different sets were used for MC and QMC. Although this is to our disadvantage, it highlights a benefit of the proxy which is to reduce unnecessary evaluations of potentially costly integrands. 
Low-discrepancy sampling appears to perform better than random sampling for training.

\hdg{Gaussian Processes} Despite their success as surrogates, GPs cope poorly with discontinuous functions and scale slowly in $N$. We compared errors of Q-NETs with Bayesian Quadrature (BQ) using the EmuKit library~\cite{emukit2019} on a step function in 1D over 100 repetitions. BQ performed $0.25\times$ better (lower error) at 16 samples but $3\times$ worse at $512$. Q-NETs are faster than BQ by factors of $2\times$ (16 samples) and $400\times$ (512 samples) per rep. Discontinuities in higher dimensional functions pose a greater challenge to GPs than to neural proxies.

\hdg{Limitations}
When integrating discontinuous functions, excessively large $k$ is counterproductive due to overfitting/ringing. This known limitation of shallow sigmoidal approximators could be addressed by bounding width~\cite{fan2018slim} and adding layers. Although, in theory, polylogarithms lend themselves to successive integration we leave the extension to multiple hidden layers as future work. 

%% file: deriv.tex
We start with the simple case when $d=1$ and $k=1$, then add neurons before developing intuition for $d=2$ and generalizing the result to the $d$-dimensional case.
When $d=1$, the matrix in equation~\ref{eq:ffnet} is reduced to a vector $\wav$ ($k\times 1$).
When $k=1$ (single neuron) parameters for \fw\  reduce to scalars $ w_1,w_2, b_1, b_2 \in \RR$ since
$$\fw(x)= \frac {w_2} {1+e^{-(w_1 x + b_1)}} + b_2 , \;\;\;\;\;\; x\in[-1,1] .$$
This can be integrated analytically to obtain
\begin{align}
\nonumber
		\mu_{1,1}(\w) & \equiv  \DefInt {-1}{1}{\fw(x)}{x} \\
\nonumber
\nonumber
		&=  
				\frac {w_2} {w_1} \left( \Sp{(w_1+b_1)} - \Sp{(-w_1+b_1)} \right) + 2b_2 
\end{align}
which consists of \verb+softplus+ terms $\Sp(x)\equiv\ln{(1+e^x)}$. Since the output layer is a linear combination of the activations of the neurons in the hidden layer, when $k>1$  the rhs is the weighted sum: 
\begin{align}
\label{eq:muk1d}
\mu_{1,k}(\w)  &=  
			 \sum\limits_{i=1}^{k}  \frac {\wbv^i} {\wav^i} 
			 		\left[\Sp{({\wav^i+\mathbf{b}_1^i})} - 		\Sp{({-\wav^i+\mathbf{b}_1^i})} \right] + 2b_2, 
\end{align}
where superscript $i$ is used to denote the $i^{th}$ element of a vector. 

 When $d=2$ and $k=1$,  the approximator network can be written as $\fw = w_2 \sig(\w_1 \x+b_1) + b_2$ where $\x = \left[x_1, x_2\right]^T$, $w_2, b_1, b_2\in \RR$ and \wav\ is a $1\times 2$ vector. Proceeding similarly to the 1D case to first integrate over one variable $x_1$ (or equally $x_2$) yields the intermediate formula, a function of $x_2$ (resp. $x_1$), which then requires a second integration to obtain $\mu_{2,1}(\w)$. The second integral is
\begin{equation}
\nonumber	
                    \DefInt {-1}{1} 
                    {\left( \frac  {w_2} {\w_1^1}  \left[ \Sp ( \w_1^1 +\w_1^2 x_2+b_1) - \Sp(-\w_1^1 + \w_1^2 x_2 + b_1) \right] + 2b_2\right)}
                    {x_2} 
\end{equation}
which can be solved to yield
\begin{align}
\nonumber
\MoveEqLeft
                \mu_{2,1}(\w) = \; 4 w_2 \; + \; 4b_2   + \; \frac  {w_2} {\w_1^1 \w_1^2} \psi \; \mathrm{where}& \\
\nonumber	
            & \psi \equiv  \left[ \Pl 2( \exp (\w_1^1 +\w_1^2  - b_1)) - \Pl 2(\exp(\w_1^1 - \w_1^2  - b_1)) \right. \\
\nonumber	
            &    \quad \left. - \; \Pl 2( \exp (-\w_1^1 + \w_1^2  -  b_1)) + \Pl 2( \exp (- \w_1^1  -\w_1^2  - b_1))  \right].
\end{align}
Here $\Pl d (x)$ represents the polylogarithm~\cite{lewin1981polylogarithms} function of order $2$ and is real when $x$ and $s$ are real. An important property of this function is that $\Pl {d+1}(x) = \int_0^x {\Pl d(t) /t}{\; \mathrm{d}t}$, which leads to the general formula for $d$-dimensional integrals with $k=1$ neurons:
\begin{align}
\nonumber
    \MoveEqLeft \mu_{d,1}(\w) =   
                    {w_2}  \sum\limits_{m=1}^{2^d} 
                    \frac 
			{(-1)^{\alpha_m}
			\; \Pl d \left( -\exp (\p^{m,.}\; \wav^\top  - b_1)\right)}
		{\prod\limits_{j=1}^{d}\w_1^j }  \\
\nonumber		
                & \quad \quad  + \; 2^d w_2 \; + \; 2^d \; b_2
\end{align}
where the rows of $2^d \times d$ matrix $\p$, where $\p^{l,m} \in \{ -1,1 \}$,  represent all $2^d$ combinations of $d$ signs, one for each element in the row vector  \wav. The contribution of each $\Pl d$ term is either positive ($\alpha_m$ is even) or negative ($\alpha_m$ is odd) depending on  whether there are an even (resp. odd) number  of $-1$s in the row $\p^{l,:}$. The case for $k$ neurons $\mu_{d,k}(\w)$ is a weighted sum of the outputs of the $k$ neurons ($\mathbf v^i, \; i=1\cdots k$), with weights given by $\wbv$. This leads to \eqnref{qnetd}.